\documentclass[aos,MSNbibl,seceqn,dvips]{arximspdf}
\usepackage{algorithm,algorithmic}
\usepackage{mathrsfs}

\doi{10.1214/13-AOS1112} 
\volume{41}
\issue{3}
\pubyear{2013}
\firstpage{1406}
\lastpage{1430}

\makeatletter

\newtheorem{theorem}{Theorem}[section]
\newtheorem{lemma}[theorem]{Lemma}
\newtheorem{proposition}[theorem]{Proposition}

\newproclaim{definition}[theorem]{Definition}
\newproclaim{remark}{Remark}

\newcommand{\arginf}{\mathop{\arg\inf}}
\renewcommand{\tilde}{\widetilde}

\renewcommand{\Pr}{\mathbb{P}}

\makeatother

\begin{document}
\begin{frontmatter}

\title{Universally consistent vertex classification for latent
positions graphs\thanksref{T1}}
\runtitle{Universally consistent vertex classification}

\thankstext{T1}{Supported in part by National Security
Science and Engineering Faculty Fellowship (NSSEFF), Johns Hopkins
University Human Language Technology Center of Excellence (JHU HLT
COE) and the XDATA program of the Defense Advanced Research Projects
Agency (DARPA) administered through Air Force Research Laboratory
contract FA8750-12-2-0303.}

\begin{aug}
\author[A]{\fnms{Minh} \snm{Tang}\ead[label=e1]{mtang10@jhu.edu}},
\author[A]{\fnms{Daniel L.} \snm{Sussman}\ead[label=e2]{dsussma3@jhu.edu}}
\and
\author[A]{\fnms{Carey E.} \snm{Priebe}\corref{}\ead[label=e3]{cep@jhu.edu}}
\runauthor{M. Tang, D. L. Sussman and C. E. Priebe}
\affiliation{Johns Hopkins University}
\address[A]{Department of Applied Mathematics\\
\quad and Statistics\\
Johns Hopkins University\\
3400 N. Charles St\\
Baltimore, Maryland 21218\\
USA\\
\printead{e1}\\
\hphantom{E-mail: }\printead*{e2}\\
\hphantom{E-mail: }\printead*{e3}} 
\end{aug}

\received{\smonth{12} \syear{2012}}
\revised{\smonth{3} \syear{2013}}

%
\begin{abstract}
In this work we show that, using the eigen-decomposition of the
adjacency matrix, we can consistently estimate feature maps
for latent position graphs with positive definite link function
$\kappa$, provided that the latent positions are i.i.d. from some
distribution $F$. We then consider the exploitation task of
vertex classification where the link function $\kappa$ belongs to the
class of universal kernels and class labels are observed for a
number of vertices tending to infinity and that the remaining
vertices are to be classified. We show that minimization of the
empirical $\varphi$-risk for some convex surrogate
$\varphi$ of 0--1 loss over a class of linear classifiers with
increasing complexities yields a universally consistent classifier,
that is, a classification rule with error converging to Bayes optimal
for any distribution $F$.
\end{abstract}

%

\end{frontmatter}

\section{Introduction}
\label{secintroduction}
The classical statistical pattern recognition setting involves
\[
(X,Y), (X_1, Y_1), (X_2, Y_2),\ldots, (X_n, Y_n) \stackrel{\mathrm
{i.i.d.}} {\sim}
F_{\mathcal{X}, \mathcal{Y}},
\]
where the $X_i \in\mathcal{X} \subset\mathbb{R}^{d}$ are observed
feature vectors, and the $Y_{i} \in\mathcal{Y} = \{-1,1\}$ are observed
class labels, for some probability distribution
$F_{\mathcal{X},\mathcal{Y}}$ on $\mathcal{X} \times\mathcal{Y}$. Let
$\mathcal{D} = \{(X_i,Y_i)\}_{i=1}^{n}$. A classifier $h(\cdot;
\mathcal{D}) \dvtx\mathcal{X} \mapsto\{-1,1\}$ whose probability of
error $\mathbb{P}[h(X; \mathcal{D}) \neq Y | \mathcal{D}]$ approaches
Bayes-optimal as $n \rightarrow\infty$ for all distributions
$F_{\mathcal{X}, \mathcal{Y}}$ is said to be \textit{universally
consistent}. For example, the k-NN classifier with $k
\rightarrow\infty, k/n \rightarrow0$ is universally consistent
\cite{stone1977consistent}.

\begin{algorithm}
\caption{Vertex classifier on graphs}
\label{algmain}
\begin{algorithmic}
\STATE\textbf{Input}: $\mathbf{A}\in
\{0,1\}^{n\times n}$, training set $\mathcal{T} \subset[n] =
\{1,2,\ldots,n\}$ and labels $\mathbf{Y}_{\mathcal{T}} = \{Y_i \dvtx
i \in\mathcal{T}\}$.
\STATE\textbf{Output}: Class labels $\{ \hat{Y}_{j} \dvtx j \in[n]
\setminus
\mathcal{T}\}$.
\STATE\textit{Step 1}: Compute the
eigen-decomposition of $\mathbf{A} =
\mathbf{U}\mathbf{S}\mathbf{U}^T$.
\STATE\textit{Step 2}: Let $d$ be the ``elbow'' in
the scree plot of $\mathbf{A}$, $\mathbf{S}_{\mathbf{A}}$ the
diagonal matrix of the top $d$ eigenvalues of $\mathbf{A}$ and
$\mathbf{U}_{\mathbf{A}}$ the corresponding columns of
$\mathbf{U}$.
\STATE\textit{Step 3}: Define
$\mathbf{Z}$ to be
$\mathbf{U}_{\mathbf{A}}
\mathbf{S}_{\mathbf{A}}^{1/2}$. Denote by $Z_i$ the $i$th row of
$\mathbf{Z}$. Define $\mathbf{Z}_{\mathcal{T}}$
the rows of $\mathbf{Z}$ corresponding to the indices set
$\mathcal{T}$. $\mathbf{Z}$ is called the adjacency spectral
embedding of $\mathbf{A}$.
\STATE
\textit{Step 4}: Find a \emph{linear} classifier
$\tilde{g}_n$ that minimizes the empirical
$\varphi$-loss when trained on $(\mathbf{Z}_{\mathcal{T}},
\mathbf{Y}_{\mathcal{T}})$ where $\varphi$ is a \emph{convex} loss
function that is a surrogate for 0--1 loss.
\STATE\textit{Step 5}: Apply $\tilde{g}_n$ on the $\{ Z_j \dvtx j
\in[n]
\setminus\mathcal{T}\}$ to obtain the $\{\hat{Y}_j \dvtx j \in
[n] \setminus\mathcal{T}\}$.
\end{algorithmic}
\end{algorithm}

In this paper, we consider the case wherein the
feature vectors are unobserved, and we observe instead a latent
position graph $G = G(X, X_1,\ldots, X_n)$ on $n + 1$ vertices with
positive definite link function $\kappa\dvtx\mathcal{X} \times
\mathcal{X} \mapsto
[0,1]$. The graph $G$ is constructed such that there is a one-to-one
relationship between the vertices of $G$ and the feature vectors $X,
X_1,\ldots, X_n$, and the edges of $G$ are \emph{conditionally
independent} Bernoulli random variables given the \emph{latent} $X,
X_1,\ldots, X_n$. We show that there exists a universally consistent
classification rule for this extension of the classical pattern
recognition setup to latent position graph models, provided that the
link function $\kappa$ is an element of the class of \emph{universal
kernels}. In particular, we show that a classifier similar to the one
described in Algorithm~\ref{algmain} is universally
consistent. Algorithm~\ref{algmain} is an example of a procedure that first
embeds data into some Euclidean space and then performs inference in that
space. These kind of procedures are popular in analyzing graph data, as
is evident from the vast literature on multidimensional scaling,
manifold learning and spectral clustering.

The above setting of classification for latent position graphs, with
$\kappa$ being the inner product in $\mathbb{R}^{d}$, was previously
considered in~\cite{sussman12univer}. It was shown there that the
eigen-decomposition of the adjacency matrix $\mathbf{A}$ yields a
consistent estimator, up to some orthogonal transformation, of the
latent vectors $X, X_1,\ldots, X_n$. Therefore, the k-NN classifier,
using the estimated vectors, with $k \rightarrow\infty, k/n
\rightarrow0$ is universally consistent. When $\kappa$ is a general,
possibly unknown link function, we cannot expect to recover the
latent vectors. However, we can obtain a consistent
estimator of some feature map $\Phi\dvtx\mathcal{X} \mapsto
\mathscr{H}$ of $\kappa$. Classifiers that use only the feature map
$\Phi$ are universally consistent if the space $\mathscr{H}$ is
isomorphic to some dense subspace of the space of measurable functions
on $\mathcal{X}$. The notion of a universal kernel
\mbox{\cite
{steinwart01supporvectormachin,sriperumbudur11univercharackernelrkhsembedmeasur,micchelli06univer}}
characterizes those $\kappa$ whose feature maps $\Phi$ induce a dense
subspace of the space of measurable functions on $\mathcal{X}$.

The structure of our paper is as follows. We introduce the framework
of latent position graphs in Section~\ref{secframework}. In
Section~\ref{secestim-feat-maps}, we show that the
eigen-decomposition of
the adjacency matrix $\mathbf{A}$ yields a consistent estimator for
a feature map $\Phi\dvtx\mathcal{X} \mapsto l_2$ of $\kappa$. We
discuss the notion of universal kernels and the problem of vertex
classification using the estimates of the feature map $\Phi$ in
Section~\ref{seccons-vert-class}. In particular, we show that the
classification rule obtained by minimizing a convex surrogate of the
0--1 loss over a class of linear classifiers in $\mathbb{R}^{d}$ is
universally consistent, provided that $d \rightarrow\infty$ in a
specified manner. We conclude the paper with a discussion of how some of
the results presented herein can be extended and other implications.

We make a brief comment on the setup of the paper. The main
contribution of the paper is the derivation of the estimated feature
maps and their use in constructing a universally consistent vertex
classifier. We have thus considered a less general setup of compact
metric spaces, linear classifiers, and convex, differentiable loss
functions.
It is possible to extends the results herein to a more general setup where the
latent positions are elements of a (non-compact) metric space, the
class of classifiers are uniformly locally-Lipschitz, and the convex
loss function satisfies the classification-calibrated property
\cite{bartlett06convex}.
%

\section{Framework}
\label{secframework}

Let $(\mathcal{X},d)$ be a compact metric space and $F$ a probability
measure on the Borel $\sigma$-field of $\mathcal{X}$. Let $\kappa
\dvtx
\mathcal{X} \times\mathcal{X} \mapsto[0,1]$ be a continuous,
positive definite kernel. Let $L^{2}(\mathcal{X}, F)$ be the space
of square-integrable functions with respect to $F$. We can
define an integral operator $\mathscr{K} \dvtx L^{2}(\mathcal{X},
F) \mapsto L^{2}(\mathcal{X}, F)$ by
\[
\mathscr{K}f(x) = \int_{\mathcal{X}} \kappa
\bigl(x,x'\bigr) f\bigl(x'\bigr) F\bigl(
\mathrm{d}x'\bigr).
\]
$\mathscr{K}$ is a compact operator and is of trace class.

Let $\{\lambda_j \}$ be the set of eigenvalues of $\mathscr{K}$
ordered as $\lambda_1 \geq\lambda_2 \geq\cdots\geq0$. Let $\{\psi
_{j} \}$
be a set of orthonormal eigenfunctions of $\mathscr{K}$ corresponding
to the $\{\lambda_j\}$, that is,
\begin{eqnarray*}
\mathscr{K}\psi_{j} &=& \lambda_{j} \psi_{j},
\\
\int_{\mathcal{X}} \psi_{i}(x) \psi_{j}(x)
\,dF(x) &=& \delta_{ij}.
\end{eqnarray*}
The following Mercer representation theorem
\cite{steinwart12mercer,cucker12} provides a representation for
$\kappa$ in terms of the eigenvalues and eigenfunctions of
$\mathscr{K}$ defined above.
%
%
\begin{theorem}
\label{thm8}
Let $(\mathcal{X},d)$ be a compact metric space and $\kappa\dvtx
\mathcal{X} \times\mathcal{X} \mapsto[0,1]$ be a continuous
positive definite kernel. Let $\lambda_1 \geq\lambda_2 \geq\cdots
\geq0$\vadjust{\goodbreak}
be the eigenvalues of $\mathscr{K}$ and $\psi_{1}, \psi_{2}, \ldots$
be the associated eigenvectors. Then
%
%
\begin{equation}
\label{eq8} \kappa\bigl(x,x'\bigr) = \sum
_{j=1}^{\infty} \lambda_{j}
\psi_{j}(x) \psi_{j}\bigl(x'\bigr).
\end{equation}
The sum in equation (\ref{eq8}) converges absolutely for each $x$
and $x'$ in $\operatorname{supp}(F) \times\operatorname{supp}(F)$ and
uniformly on $\operatorname{supp}(F) \times\operatorname{supp}(F)$. Let
$\mathscr{H}$ denote the reproducing kernel Hilbert space of
$\kappa$. Then the elements $\eta\in\mathscr{H}$ are of the form
%
%
\begin{equation}
\label{eq82} \eta= \sum_{j} a_{j}
\sqrt{\lambda}_{j} \psi_{j}\qquad\mbox{with $(a_j)
\in l_2$},
\end{equation}
and the inner product on $\mathscr{H}$ is given by
%
%
\begin{equation}
\label{eq83} \biggl\langle\sum_{j}
a_{j} \sqrt{\lambda}_{j} \psi_{j}, \sum
_{j} b_{j} \sqrt{\lambda}_{j}
\psi_{j} \biggr\rangle_{\mathscr
{H}} = \sum
_{j} a_{j} b_{j}.
\end{equation}
\end{theorem}
By Mercer's representation theorem, we have $\kappa(\cdot,x) =
\sum_{j} \sqrt{\lambda}_{j} \psi_{j}(x) \*\sqrt{\lambda}_{j} \psi
_{j}(\cdot)$. We thus define the
feature map $\Phi\dvtx\mathcal{X} \mapsto l_{2}$ by
\[
\Phi(x) = \bigl(\sqrt{\lambda_{j}} \psi_{j}(x) \dvtx j =
1, 2, \ldots\bigr).
\]
Let $d$ be an integer with $d \geq1$. We also define the following
map $\Phi_{d} \dvtx\mathcal{X} \mapsto\mathbb{R}^{d}$
\[
\Phi_{d}(x) = \bigl(\sqrt{\lambda}_{j}
\psi_{j}(x) \dvtx j = 1, 2,\ldots, d\bigr).
\]
We will refer to $\Phi_{d}$ as the truncation of $\Phi$ to
$\mathbb{R}^{d}$.

Now, for a given $n$, let $X_1,\ldots, X_n
\stackrel{\mathrm{i.i.d.}}{\sim} F$. Define $\mathbf{K} =
(\kappa(X_i,X_j))_{i,j=1}^{n}$. Let $\mathbf{A}$ be a symmetric
random hollow matrix where the entries $\{\mathbf{A}_{ij}\}_{i < j}$
are \emph{conditionally independent} Bernoulli random variables given
the $\{X_i\}_{i=1}^{n}$ with
$\mathbb{P}[\mathbf{A}_{ij} = 1] = \mathbf{K}_{ij}$ for all $i,j \in
[n]$, $i < j$. $\mathbf{A}$ is the adjacency matrix corresponding to a
graph with vertex set $\{1,2,\ldots,n\}$. A graph $G$ whose adjacency
matrix $\mathbf{A}$ is constructed as above is an instance of a latent
position graph~\cite{Hoff2002} where the latent positions are sampled
according to $F$, and the link function is $\kappa$.

\subsection{Related work}
The latent position graph model and the related latent space
approach~\cite{Hoff2002} is widely used in network
analysis. It is a generalization of the stochastic block model (SBM)
\cite{Holland1983} and variants such as the degree-corrected SBM
\cite{karrer11stoch} or the mixed-membership SBM~\cite{Airoldi2008}
or the random dot product graph model~\cite{young2007random}. It is
also closely related to the inhomogeneous random graph model
\cite{bollobas07} or the exchangeable graph model
\cite{diaconis08graphlimitexchanrandomgraph}.

There are two main sources of randomness in latent position graphs. The
first source of randomness is due to the sampling procedure, and the
second source of randomness is due to the conditionally independent
Bernoulli trials that gave rise to the edges of the graphs.
The randomness in the sampling procedure and its effects on spectral
clustering and/or kernel PCA have been widely studied. In the manifold
learning literature, the latent positions are sampled from some
manifold in Euclidean space and
\cite{belkin05towarlaplac,hein07converlaplac,hein5fromlaplac}
among others studied the convergence of the various graph Laplacian
matrices to their corresponding Laplace--Beltrami operators on the
manifold. The authors of \cite
{luxburg08consis,rosasco10integoperat} studied the
convergence of the eigenvalues and eigenvectors of the graph Laplacian
to the eigenvalues and eigenfunctions of the corresponding operators
in the spectral clustering
setting.

The matrix $\mathbf{K}/n$ can be considered as an approximation of
$\mathscr{K}$ for large $n$; that is, we expect the eigenvalues and
eigenvectors of $\mathbf{K}/n$ to converge to the eigenvalues and
eigenfunctions of $\mathscr{K}$ in some sense. This convergence is
important in understanding the theoretical properties of kernel PCA;
see, for example,
\cite
{rosasco10integoperat,braun06accurerrorboundeigenkernelmatrix,shawe-taylor05eigengramgenerpca,bengio04learnpca,koltchinskii00random,zwald06}.
We summarized some of the results
from the literature that directly pertain to the current paper in
Appendix~\ref{secspectra-mathbfk-spec}.

The Bernoulli trials at each edge and their effects had also been
studied. For example, a result in
\cite{chatterjee12matrixuniversingulvaluethres} on matrix
estimation for noise-pertubed
and subsampled matrices showed that by thresholding the dimensions
in the singular value
decomposition of the adjacency matrix, one can recover an estimate
$\hat{\mathbf{K}}$ of the kernel matrix $\mathbf{K}$ with small $\|
\hat{\mathbf{K}} - \mathbf{K} \|$. Oliveira~\cite{oliveira2010concentration}
studied the convergence of the eigenvalues and eigenvectors of the
adjacency matrix $\mathbf{A}$ to that of the integral operator
$\mathscr{K}$ for the class of inhomogeneous random graphs. The
inhomogeneous random graphs in~\cite{oliveira2010concentration} have
latent positions that are uniform $[0,1]$ random variables with the
link function $\kappa$ being arbitrary symmetric functions.

As we have mentioned in Section~\ref{secintroduction},
Algorithm~\ref{algmain} is an example of a popular approach in
multidimensional scaling, manifold learning and spectral clustering
where inference on graphs proceeds by first embedding the graph into
Euclidean space followed by inference on the resulting embedding. It
is usually assumed that the embedding is conducive to the subsequent
inference tasks. Justification can also be provided based on the
theoretical results about convergence, for example, the convergence of the
eigenvalues and eigenvectors to the eigenvalues and eigenfunctions of
operators, or the convergence of the estimated entries, cited
above. However, these justifications do not consider the subsequent
inference problem; that is, these convergence results do not directly
imply that inference using the embeddings are meaningful. Recently, the
authors of
\cite{rohe2011spectral,sussman12,fishkind2012consistent,chaudhuri12spect}
showed that the clustering using the embeddings are meaningful, that is,
consistent, for graphs based on the stochastic block model and the
extended planted partition model. The main impetus for this paper is
to give similar theoretical justification for the
classification setting. The latent position graph model is thus a
surrogate model---a~widely-used model with sufficiently simple
structure that allows for clear, concise theoretical results.

\section{Estimation of feature maps}
\label{secestim-feat-maps}
We assume the setting of Section~\ref{secframework}. Let us denote by
$\mathcal{M}_{d}(\mathbb{R})$ and $\mathcal{M}_{d,n}(\mathbb{R})$ the
set of $d \times d$ matrices and $d \times n$ matrices\vspace*{1pt} on
$\mathbb{R}$, respectively. Let $\tilde{\mathbf{U}}_\mathbf{A}
\tilde{\mathbf{S}}_\mathbf{A} \tilde{\mathbf{U}}_\mathbf{A}^\top
$ be
the eigen-decomposition of $\mathbf{A}$. For a given $d \geq1$, let
$\mathbf{S}_\mathbf{A}\in\mathcal{M}_{d}(\mathbb{R})$ be the diagonal
matrix comprised of the $d$ largest eigenvalues of $\mathbf{A}$, and
let $\mathbf{U}_\mathbf{A}\in\mathcal{M}_{n,d}(\mathbb{R})$ be the
matrix comprised of the corresponding eigenvectors. The matrices
$\mathbf{S}_\mathbf{K}$ are $\mathbf{U}_\mathbf{K}$ are defined
similarly. For a matrix $\mathbf{M}$, $\| \mathbf{M} \|$ refers to the
spectral norm of $\mathbf{M}$ while $\| \mathbf{M} \|_{F}$ refers to
the Frobenius norm of $\mathbf{M}$. For a vector $v \in
\mathbb{R}^{n}$, $\|v\|$ will denote the Euclidean norm of $v$.

The key result of this section is the following theorem which shows
that, given that there is a gap in the spectrum of $\mathscr{K}$ at
$\lambda_{d}(\mathscr{K})$, by using the eigen-decomposition of
$\mathbf{A}$ we can accurately estimate the truncated map $\Phi_{d}$
in Section~\ref{secframework} up to an orthogonal transformation. We note
that the dependence on $F$, the distribution of the $\{X_i\}$, in the
following result is implicit in the definition of the spectral gap
$\delta_{d}$
of $\mathscr{K} \dvtx L^{2}(\mathcal{X}, F) \mapsto L^{2}(\mathcal
{X}, F)$.
%
%
\begin{theorem}
\label{thm1}
Let $d \geq1$ be given. Denote by $\delta_{d}$ the quantity
$\lambda_{d}(\mathscr{K}) - \lambda_{d+1}(\mathscr{K})$, and suppose
that $\delta_{d} > 0$. Then with probability greater than $1-2
\eta$, there exists a unitary matrix $\mathbf{W} \in
\mathcal{M}_{d}(\mathbb{R})$ such that
%
%
\begin{equation}\label{eqXBnd}
\bigl\| \mathbf{U}_\mathbf{A} \mathbf{S}_\mathbf{A}^{1/2}
\mathbf{W} - \bolds{\Phi}_{d} \bigr\|_{F} \leq27
\delta_{d}^{-2} \sqrt{d \log{(n/\eta)}},
\end{equation}
where $\bolds{\Phi}_{d}$ denotes the matrix in
$\mathcal{M}_{n,d}(\mathbb{R})$ whose $i$th row is
$\Phi_{d}(X_i)$. Let us denote by $\hat{\Phi}_{d}(X_i)$ the
$i$th row of $\mathbf{U}_\mathbf{A}\mathbf{S}_\mathbf
{A}^{1/2}\mathbf{W}$. Then,
for each $i\in[n]$ and any $\varepsilon> 0$,
%
%
\begin{equation}\label{eqxiBnd}
\Pr\bigl[\bigl\|\hat{\Phi}_{d}(X_i) -\Phi_{d}(X_i)
\bigr\| > \varepsilon\bigr] \leq27 \delta_d^{-2}
\varepsilon^{-1} \sqrt{\frac{6d \log{n}}{n}}.
\end{equation}
\end{theorem}

We now proceed to prove Theorem~\ref{thm1}. A rough sketch of the
argument goes as follows. First we will show that the projection of
$\mathbf{A}$ onto the subspace spanned by $\mathbf{U}_{\mathbf{A}}$ is
``close'' to the projection of $\mathbf{K}$ onto the subspace spanned
by $\mathbf{U}_\mathbf{K}$. Then we will use results on the convergence
of spectra of $\mathbf{K}$ to the spectra of $\mathscr{K}$ to show that
the subspace spanned by $\mathbf{U}_{\mathbf{A}}$ is also ``close'' to
the subspace spanned by~$\Phi_{d}$. We note that, for conciseness and
simplicity in the exposition, all probability statements involving the
matrix $\mathbf{A}$ or its related quantities, for example,
$\mathbf{U}_{\mathbf{A}}, \mathbf{S}_{\mathbf{A}}$, are assumed to hold
conditionally on the $\{X_1, X_2,\ldots, X_n\}$.

We need the following bound for the perturbation $\mathbf{A} -
\mathbf{K}$ from~\cite{oliveira2010concentration}. The convergence of
the spectra of $\mathbf{A}$ to that of $\mathscr{K}$ as given by
Theorem 6.1 in~\cite{oliveira2010concentration} is similar to that
given in the proof of Theorem~\ref{thm1} in the current paper, but
there are sufficient differences between the two settings, and we do
not see an obvious way to apply the conclusions of Theorem 6.1 in
\cite{oliveira2010concentration} to the current paper.
%
%
\begin{proposition}
\label{prop3}
For $\mathbf{A}$ and $\mathbf{K}$ as defined above, with probability
at least $1 - \eta$, we have
%
%
\begin{equation}
\label{eq10} \| \mathbf{A} - \mathbf{K} \| \leq2 \sqrt{ \Delta\log
{(n/\eta)}}
\leq2\sqrt{n \log{(n/\eta)}},
\end{equation}
where $\Delta$ is the maximum vertex degree.
\end{proposition}
The constant in equation (\ref{eq10}) was obtained by replacing a
concentration inequality in~\cite{oliveira2010concentration} with a
slightly stronger inequality from~\cite{tropp11freed}. We now show
that the projection matrix for the subspace spanned by
$\mathbf{U}_{\mathbf{A}}$ is close to the projection matrix for the
subspace spanned by $\mathbf{U}_{\mathbf{K}}$.
%
%
\begin{proposition}
\label{prop4}
Let $\mathcal{P}_{\mathbf{A}} = \mathbf{U}_{\mathbf{A}}
\mathbf{U}_{\mathbf{A}}^{T}$ and $\mathcal{P}_{\mathbf{K}} =
\mathbf{U}_{\mathbf{K}} \mathbf{U}_{\mathbf{K}}^{T}$. Denote by
$\delta_d$ the quantity $\lambda_{d}(\mathscr{K}) -
\lambda_{d+1}(\mathscr{K})$, and suppose that $\delta_d > 0$. If
$n$ is such that $\delta_d \geq8(1 + \sqrt{2}) n^{-1/2}
\sqrt{\log{(n/\eta)}}$. Then with probability at least $ 1 - 2
\eta$,
%
%
\begin{equation}
\label{eq4} \|\mathcal{P}_{\mathbf{A}} - \mathcal{P}_{\mathbf{K}} \|
\leq4
\sqrt{\frac{\log{(n/\eta)}}{n \delta_d^{2}}}.
\end{equation}
\end{proposition}

\begin{pf}
By equation (\ref{eq30}) in Theorem~\ref{thm5}, we have with
probability at least $1 - \eta$,
\[
\frac{\lambda_{d}(\mathbf{K})}{n} - \frac{\lambda_{d+1}(\mathbf{K})}{n}
\geq\delta_{d} - 4 \sqrt{2}
\sqrt{\frac{\log{(2/\eta)}}{n}}.
\]
Now, let $S_1$ and $S_2$ be defined as
\begin{eqnarray*}
S_1 &=& \bigl\{ \lambda\dvtx\lambda\geq n \lambda_{d}(
\mathbf{K}) - 2 \sqrt{n \log{(n/\eta)}}\bigr\},
\\
S_2 &=& \bigl\{ \lambda\dvtx\lambda< n \lambda_{d+1}(
\mathbf{K}) + 2 \sqrt{n \log{(n/\eta)}}\bigr\}.
\end{eqnarray*}
Then we have, with probability at least $1 - \eta$,
%
%
\begin{eqnarray}
\label{eq57} \operatorname{dist}(S_1, S_2) &\geq& n
\delta_d - 4 \sqrt{2} \sqrt{n \log{(2/\eta)}} - 4 \sqrt{n \log{(n/
\eta)}}
\nonumber\\[-8pt]\\[-8pt]
&\geq& n \delta_{d} - 4( 1 + \sqrt{2}) \sqrt{n \log{(n/\eta)}}.
\nonumber
\end{eqnarray}
Suppose for the moment that $S_1$ and $S_2$ are disjoint, that is, that
$\operatorname{dist}(S_1,\break S_2) > 0$. Let $\mathcal{P}_{\mathbf
{A}}(S_1)$ be
the matrix for the orthogonal projection onto the~sub\-space spanned by
the eigenvectors of $\mathbf{A}$ whose corresponding eigenvalues lies
in $S_1$. Let $\mathcal{P}_{\mathbf{K}}$ be defined similarly. Then by
the $\operatorname{sin}\bolds{\Theta}$ theorem~\cite{davis70} we have
\[
\bigl\| \mathcal{P}_{\mathbf{A}}(S_1) - \mathcal{P}_{\mathbf{K}}(S_1)
\bigr\| \leq\frac{\| \mathbf{A} - \mathbf{K} \|}{\operatorname{dist}(S_1, S_2)}.
\]
By equation (\ref{eq57}) and Proposition~\ref{prop3},
we have, with probability at least $(1 - 2 \eta)$,
\[
\bigl\| \mathcal{P}_{\mathbf{A}}(S_1) - \mathcal{P}_{\mathbf{K}}(S_1)
\bigr\| \leq\frac{ 2 \sqrt{ n \log{(n/\eta)}}}{ n \delta_{d} - 4(
1 + \sqrt{2}) \sqrt{n \log{(n/\eta)}}} \leq4 \sqrt{ \frac{
\log{(n/\eta)}}{n \delta_d^{2}}},
\]
provided that $4(1 + \sqrt{2}) \sqrt{n \log{(n/\eta)}} \leq n
\delta_d/2$.\vadjust{\goodbreak}

To complete the proof, we note that if $4(1 + \sqrt{2}) \sqrt{n
\log{(n/\eta)}} \leq n \delta_d/2$, then $S_1$ and $S_2$ are
disjoint. Thus $\mathcal{P}_{\mathbf{K}}(S_1) =
\mathbf{U}_{\mathbf{K}} \mathbf{U}_{\mathbf{K}}^{T}$. Finally, if
$\|\mathbf{A}
- \mathbf{K} \| \leq2 \sqrt{ n \log{(n/\eta)}}$, then the
eigenvalues of
$\mathbf{A}$ that lie in $S_1$ are exactly the $d$ largest
eigenvalues of $\mathbf{A}$ and $\mathcal{P}_{\mathbf{A}}(S_1) =
\mathbf{U}_{\mathbf{A}} \mathbf{U}_{\mathbf{A}}^{T}$.
Equation (\ref{eq4}) is thus established.
\end{pf}

Let $\mathscr{H}$ be the reproducing kernel Hilbert space for
$\kappa$. We now introduce a linear operator
$\mathscr{K}_{\mathscr{H},n}$ on $\mathscr{H}$ defined as follows:
\[
\mathscr{K}_{\mathscr{H},n} \eta= \frac{1}{n} \sum
_{i=1}^{n} \bigl\langle\eta, \kappa(\cdot,
{X_i}) \bigr\rangle_{\mathscr{H}} \kappa(\cdot,
{X_i}).
\]
The operator $\mathscr{K}_{\mathscr{H},n}$ is the extension of
$\mathbf{K}$ as an operator on $\mathbb{R}^{n}$ to an operator on
$\mathscr{H}$. That is, $\mathscr{K}_{\mathscr{H},n}$ is a linear
operator on $\mathscr{H}$ induced by $\kappa$ and the $X_1, X_2,\ldots,
X_n$. The eigenvalues of $\mathscr{K}_{\mathscr{H},n}$ and
$\mathbf{K}$ coincide, and furthermore, an eigenfunction of
$\mathscr{K}_{\mathscr{H},n}$ is a linear interpolation of the
corresponding eigenvector of $\mathbf{K}$. The reader is referred to
Appendix~\ref{secspectra-mathbfk-spec} for more details.

The next result states that the rows of $\mathbf{U}_{\mathbf{K}}
\mathbf{S}_{\mathbf{K}}^{1/2}$ correspond to projecting the $\Phi
(X_i)$ using
$\hat{\mathcal{P}}_{d}$, where $\hat{\mathcal{P}}_{d}$ is the
projection onto the $d$-dimensional subspace spanned by the
eigenfunctions associated with the $d$ largest
eigenvalues of $\mathscr{K}_{\mathscr{H},n}$. We note that for
large $n$, $\hat{\mathcal{P}}_{d}$ is close to the projection onto the
$d$-dimensional subspace spanned by the eigenfunctions associated with
the $d$ largest eigenvalues of
$\mathscr{K}$ with high probability; see Theorem~\ref{thm5}.

%
%
\begin{lemma}
\label{lem4}
Let $\hat{\mathcal{P}}_{d}$ be the projection onto the
subspace spanned by the eigenfunctions corresponding to the $d$
largest eigenvalues of $\mathscr{K}_{\mathscr{H},n}$. The rows of
$\mathbf{U}_{\mathbf{K}} \mathbf{S}_{\mathbf{K}}^{1/2}$ then
correspond, up to some orthogonal transformation, to projections of
the feature map $\Phi$ onto $\mathbb{R}^{d}$ via
$\hat{\mathcal{P}}_{d}$, that is, there exists a unitary matrix
$\mathbf{W} \in\mathcal{M}_{d}(\mathbb{R})$ such that
%
%
\begin{equation}
\label{eq68} \mathbf{U}_{\mathbf{K}} \mathbf{S}_{\mathbf{K}}^{1/2}
\mathbf{W} = \bigl[\imath\bigl(\hat{\mathcal{P}_{d}}\bigl(
\Phi(X_1)\bigr)\bigr)^{T} | \cdots| \imath\bigl(\hat{
\mathcal{P}_{d}}\bigl(\Phi(X_n)\bigr)\bigr)^{T}
\bigr]^{T},
\end{equation}
where $\imath$ is the isometric isomorphism of a finite-dimensional
Hilbert space onto~$\mathbb{R}^{d}$.
\end{lemma}

%
%


The proof of Lemma~\ref{lem4} is given in the \hyperref[app]{Appendix}.
\begin{pf*}{Proof of Theorem~\ref{thm1}}
%
We first note that the sum of any row of $\mathbf{A}$ is bounded
from above by $n$, thus $\| \mathbf{A} \| \leq n$. Similarly, $\|
\mathbf{K} \| \leq n$. On combining equation (\ref{eq4}) and
equation (\ref{eq10}), we have, with probability at least $1 - 2 \eta$,
\begin{eqnarray*}
\| \mathcal{P}_{\mathbf{A}} \mathbf{A} - \mathcal{P}_{\mathbf{K}}
\mathbf{K}\|
&\leq&
\bigl\| \mathcal{P}_{\mathbf{A}} (\mathbf{A} - \mathbf{K}) \bigr\| + \bigl\| (
\mathcal{P}_{\mathbf{A}} - \mathcal{P}_{\mathbf{K}}) \mathbf{K} \bigr\|
\\
&\leq&
2 \sqrt{n \log{(n/\eta)}} + 4 \delta_d^{-1} \sqrt{
n \log{(n/\eta)}}
\\
&\leq&
6 \delta_{d}^{-1} \sqrt{n \log{(n/\eta)}}.
\end{eqnarray*}
%
By Lemma~\ref{lem3} in the \hyperref[app]{Appendix}, there exists an orthogonal
$\mathbf{W} \in\mathcal{M}_{d}(\mathbb{R})$ such that
\begin{eqnarray*}
\bigl\| \mathbf{U}_{\mathbf{A}} \mathbf{S}_{\mathbf{A}}^{1/2} \mathbf{W}
- \mathbf{U}_{\mathbf{K}} \mathbf{S}_{\mathbf{K}}^{1/2} \bigr\|
&\leq&
6\delta_{d}^{-1} \sqrt{n \log{(n/\eta)}} \frac{ \sqrt{d
\| \mathcal{P}_{\mathbf{A}} \mathbf{A} \|} + \sqrt{d \|
\mathcal{P}_{\mathbf{K}} \mathbf{K} \|}}{\lambda_{d}(\mathbf{K})}
\\
&\leq&
12 \delta_{d}^{-1} \frac{n \sqrt{\log{(n/\eta)}}}{\lambda
_{d}(\mathbf{K})}.
\end{eqnarray*}
We note that $\lambda_{d}(\mathbf{K}) \geq n \lambda_{d}(\mathscr{K})/2$
provided that $n$ satisfies $\lambda_{d}(\mathscr{K}) > 4 \sqrt{2}\*
\sqrt{n^{-1} \log{(n/\eta)}}$. Thus, we have
%
%
\begin{equation}
\label{eq48}\qquad \bigl\| \mathbf{U}_{\mathbf{A}} \mathbf{S}_{\mathbf{A}}^{1/2}
\mathbf{W} - \mathbf{U}_{\mathbf{K}} \mathbf{S}_{\mathbf{K}}^{1/2}
\bigr\|_{F} \leq24 \delta_{d}^{-1} \frac{ \sqrt{d \log{(n/\eta)}}}{
\lambda_{d}(\mathscr{K})}
\leq24 \delta_{d}^{-2} \sqrt{d \log{(n/\eta)}}
\end{equation}
with probability at least $1 - 2 \eta$.

Now, by Lemma~\ref{lem4}, the rows of $\mathbf{U}_{\mathbf{K}}
\mathbf{S}_{\mathbf{K}}^{1/2}$ are (up to
some orthogonal transformation) the
projections of the feature map $\Phi$ onto $\mathbb{R}^{d}$ via
$\hat{\mathcal{P}}_{d}$. On the other hand, $\Phi_{d}(X)$ is the
projection of $\kappa(\cdot, X)$ onto $\mathbb{R}^{d}$ via $\mathcal
{P}_{d}$. By
Theorem~\ref{thm5} in the \hyperref[app]{Appendix}, for all $X$, we have
\[
\bigl\|\hat{\mathcal{P}}_{d} \kappa(\cdot, X) - \mathcal{P}_{d}
\kappa(\cdot, X) \bigr\|_{\mathscr{H}} \leq\| \hat{\mathcal{P}}_{d} -
\mathcal{P}_{d} \|_{\mathrm{HS}} \bigl\| \kappa(\cdot, X) \bigr\|_{\mathscr{H}}
\leq2\sqrt{2}\frac{\sqrt{\log(1/\eta)}}{\delta_{d} \sqrt{n}}
\]
with probability at least $1 - 2\eta$. We therefore have, for some
orthogonal $\tilde{\mathbf{W}} \in
\mathcal{M}_{d}(\mathbb{R})$,
%
%
\begin{equation}
\label{eq74} \bigl\| \mathbf{U}_{\mathbf{K}} \mathbf{S}_{\mathbf{K}}^{1/2}
\tilde{\mathbf{W}} - \bolds{\Phi}_{d} \bigr\|_{F} \leq2\sqrt{2}
\frac{\sqrt{\log{(1/\eta)}}}{\delta_d}
\end{equation}
with probability at least $1 - 2\eta$. Equation (\ref{eqXBnd}) in the
statement of the theorem then follows from equation (\ref{eq48})
and equation (\ref{eq74}).

To show equation (\ref{eqxiBnd}), we first note that as the
$\{X_i\}_{i=1}^{n}$ are independent and identically distributed, the
$\{\hat{\Phi}_{d}(X_i)\}_{i=1}^{n}$ are exchangeable and hence identically
distributed. Let $\eta= n^{-2}$. By conditioning on the event in
equation (\ref{eqXBnd}), we have
%
%
\begin{eqnarray}
\label{eq19} \mathbb{E}\bigl[\bigl\| \hat{\Phi}_{d}(X_i) -
\Phi_{d}(X_i) \bigr\|\bigr] &\leq&
\sqrt{\mathbb{E}\bigl[\bigl\|
\hat{\Phi}_{d}(X_i) - \Phi_{d}(X_i)
\bigr\|^{2}\bigr]}
\nonumber
\\
&\leq& \sqrt{ \frac{1}{n} \mathbb{E}\bigl[\|\hat{\bolds{
\Phi}}_{d}- \bolds{\Phi}_{d} \|_F^2
\bigr]}
\nonumber\\[-8pt]\\[-8pt]
&\leq& \frac{1}{\sqrt{n}}\sqrt{ \biggl(1-\frac{2}{n^2} \biggr)
\bigl(27 \delta_d^{-2} \sqrt{3d \log{n}}
\bigr)^{2} + \frac{2}{n^2} 2n}
\nonumber
\\
&\leq& 27 \delta_{d}^{-2} \sqrt{ \frac{6 d \log{n}}{n}},
\nonumber
\end{eqnarray}
because the worst case bound is $\|\hat{\bolds{\Phi}}_{d}
-\bolds{\Phi}_{d} \|_{F} \leq2n$ with probability $1$.
Equation (\ref{eqxiBnd}) follows from equation (\ref{eq19}) and
Markov's inequality.
\end{pf*}

\section{Universally consistent vertex classification}
\label{seccons-vert-class} The results in
Section~\ref{secestim-feat-maps} show that by using the
eigen-decomposition of $\mathbf{A}$, we can consistently estimate the
truncated feature map $\Phi_{d}$ for any fixed, finite $d$ (up to an
orthogonal transformation). In the subsequent discussion, we will
often refer to the rows of the eigen-decomposition of $\mathbf{A}$,
that is, the rows of $\mathbf{U}_{\mathbf{A}} \mathbf{S}_{A}^{1/2}$ as
the \emph{estimated vectors}. Sussman, Tang and Priebe \cite
{sussman12univer} showed that,
for the dot product kernel on a finite-dimensional space
$\mathcal{X}$, the $k$-nearest-neighbors classifier on
$\mathbb{R}^{d}$ is universally consistent when we select the
neighbors using the estimated vectors rather than the true but unknown
latent positions. This result can be trivially extended to the setting
for an arbitrary finite-rank kernel $\kappa$ as long as the feature
map $\Phi$ of $\kappa$ is injective. It is also easy to see that if
the feature map $\Phi$ is not injective, then any classifier that uses
only the estimated vectors (or the feature map $\Phi$) is no longer
universally consistent. This section is concerned with the setting
where the kernel $\kappa$ is an infinite-rank kernel with an injective
feature map $\Phi$ onto $l_2$. Well-known examples of these kernels
are the class of universal kernels
\cite
{sriperumbudur11univercharackernelrkhsembedmeasur,micchelli06univer,steinwart01supporvectormachin}.
%
%
\begin{definition}
\label{def1}
A continuous kernel $\kappa$ on some metric space $(\mathcal{X},d)$
is a universal kernel if for some feature map $\Phi\dvtx
\mathcal{X} \mapsto H$ of $\kappa$ to some Hilbert space~$H$, the
class of
functions of the form
\[
\mathscr{F}_{\Phi} = \bigl\{ \langle w, \Phi\rangle_{H} \dvtx
w \in H \bigr\}
\]
is dense in $\mathscr{C}(\mathcal{X})$; that is, for
any continuous function $g \dvtx\mathcal{X} \mapsto\mathbb{R}$
and any $\varepsilon> 0$, there
exists a $f \in\mathscr{F}_{\Phi}$ such that $\|f - g \|_{\infty} <
\varepsilon$.
\end{definition}
We note that if $\mathscr{F}_{\Phi}$ is dense in $\mathscr
{C}(\mathcal{X})$
for some feature map $\Phi$ and $\Phi' \dvtx\mathcal{X} \mapsto
H'$ is
another feature map of $\kappa$, then $\mathscr{F}_{\Phi'}$ is also dense
in $\mathscr{C}(\mathcal{X})$, that is, the universality of $\kappa$ is
independent of the choice for its feature map. Furthermore, every
feature map of a universal kernel is injective.

The following result lists several well-known universal kernels.
%
%
\begin{proposition}[(\cite{steinwart01supporvectormachin,micchelli06univer})]
\label{prop2}
Let $S$ be a compact subset of $\mathbb{R}^{d}$. Then the following
kernels are universal on $S$:
\begin{itemize}
\item the exponential kernel $\kappa(x,y) = \exp(\langle x, y \rangle)$;
\item the Gaussian kernel $\kappa(x,y) = \exp( - \| x - y
\|^{2}/\sigma^{2})$ for all $\sigma> 0$;
\item the binomial kernel $\kappa(x,y) = (1 - \langle x, y
\rangle)^{-\alpha}$ for $\alpha> 0$;
\item the inverse multiquadrics $\kappa(x,y) = (c^{2} + \| x
- y \|^{2})^{-\beta}$ with $c > 0$ and $\beta> 0$.
\end{itemize}
\end{proposition}

If the kernel matrix $\mathbf{K}$ is known, then results on the
universal consistency of support vector machines with universal
kernels are available; see, for example,~\cite{steinwart02suppor}.
If the
feature map $\Phi$ is known, then Biau, Bunea\vadjust{\goodbreak} and Wegkamp~\cite
{biau05functhilberspaces}
showed that the $k$-nearest-neighbors on $\Phi_{d}$ are universally
consistent as $k \rightarrow\infty$ and $d \rightarrow\infty$ where
$k$ and $d$ are chosen using a structural risk minimization approach.

Our universally consistent classifier operates on the estimated
vectors and is based on an empirical risk minimization
approach. Namely, we will show that the classifier that minimizes a
convex surrogate $\varphi$ for 0--1 loss from a class of linear
classifiers $\mathcal{C}^{(d_n)}$ is universally consistent provided
that the convex surrogate $\varphi$ satisfies some mild conditions and
that the complexity of the class $\mathcal{C}^{(d_n)}$ grows in a
controlled manner.

First, we will expand our framework to the classification setting. Let
$\mathcal{X}$ be as in Section~\ref{secframework}, and let
$F_{\mathcal{X},Y}$ be a distribution on $\mathcal{X}
\times\{-1,1\}$. Let $(X_1,Y_1),\ldots,
(X_{n+1}, Y_{n+1})\stackrel{\mathrm{i.i.d.}}{\sim} F_{\mathcal
{X},Y}$, and let
$\mathbf{K}$ and $\mathbf{A}$ be as in Section~\ref{secframework}. The
$\{Y_i\}$ are the class labels for the vertices in the graph
corresponding to the adjacency matrix $\mathbf{A}$.

We suppose that we observe only $\mathbf{A}$, the adjacency matrix,
and $Y_1,\ldots,Y_n$, the class labels for all but the last vertex.
Our goal is to accurately classify this last vertex, so for
convenience of notation we shall define $X:=X_{n+1}$ and $Y:=Y_{n+1}$.
Let the
rows of $\mathbf{U}_\mathbf{A}\mathbf{S}_\mathbf{A}^{1/2}$ be denoted
by $\zeta_{d}(X_1),\ldots,\zeta_{d}(X_{n+1})$ (even though the $X_i$
are unobserved/unknown). We want to find a classifier
$h_n$ such that, for any distribution $F_{\mathcal{X},Y}$,
\begin{eqnarray*}
\mathbb{E}[L_n]&:=&\mathbb{E}\bigl[\Pr\bigl[h_n\bigl(
\zeta_{d_n}(X)\bigr) \neq Y | \bigl(\zeta_{d_n}(X_1),Y_1
\bigr),\ldots,\bigl(\zeta_{d_n}(X_n),Y_n\bigr)
\bigr]\bigr] \\
&\to&\Pr\bigl[h^*(X)\neq Y\bigr]=:L^*,
\end{eqnarray*}
where $h^{*}$ is the Bayes-optimal classifier, and $L^{*}$ is its
associated Bayes-risk.

Let $\mathcal{C}^{(d)}$ be the class of linear classifiers using the
truncated feature map $\Phi_{d}$ whose linear coefficients are
normalized to have norm at most $d$, that is, $g \in\mathcal
{C}^{(d)}$, if
and only if $g$ is of the form
%
%
\begin{equation}
\label{eq11} g(x) = \cases{ 1, &\quad if $\bigl\langle w, \Phi_{d}(x)
\bigr\rangle> 0$,
\vspace*{1pt}\cr
-1, &\quad if $\bigl\langle w, \Phi_{d}(x) \bigr\rangle
\leq0$,}
\end{equation}
for some $w \in\mathbb{R}^{d}$ with $\|w\| \leq d$. We note that the
$\{\mathcal{C}^{(d)}\}$ are increasing, that is, $\mathcal{C}^{(d)}
\subset\mathcal{C}^{(d')}$ for $d < d'$ and that $\bigcup_{d \geq1}
\mathcal{C}^{(d)} = \mathscr{F}_{\Phi} = \{ \langle w,
\Phi\rangle_{\mathscr{H}} \dvtx w \in\mathscr{H}\}$. Because $\kappa$
is universal, $\mathcal{F}_{\Phi}$ is dense in
$\mathcal{C}(\mathcal{X})$ and as $\mathcal{X}$ is compact,
$\mathcal{F}_{\Phi}$ is dense in the space of measurable functions on
$\mathcal{X}$. Thus $\lim_{d \rightarrow\infty} \inf_{g
\in\mathcal{C}^{(d)}} L(g) = L^{*}$ and so one can show that empirical
risk minimization over the class $\mathcal{C}^{(d_n)}$ for any
increasing and divergent sequence $(d_n)$ yields a universally
consistent classifier (Theorem 18.1
in~\cite{devroye1996probabilistic}). The remaining part of this section
is concerned with modifying this result so that it applies to the
estimated feature map $\zeta_{d}$ instead of the true feature
map~$\Phi_{d}$.
We now describe a setup for empirical risk minimization over
$\mathcal{C}^{(d)}$ for increasing $d$ where we use
the estimated $\zeta_{d}$ in place of the $\Phi_{d}$. Let us
write $\hat{L}_{n}(w; \zeta_{d})$ for the empirical error
when using the $\zeta_{d}$, that is,
\[
\hat{L}_{n}(w; \zeta_{d}) = \frac{1}{n} \sum
_{i=1}^{n} \mathbf{1}\bigl\{\operatorname{sign}\bigl(\bigl
\langle w, \zeta_{d}(X_i) \bigr\rangle\bigr) \neq
Y_i\bigr\} \leq\frac{1}{n} \sum_{i=1}^{n}
\mathbf{1} \bigl\{ Y_i \bigl\langle w, \zeta_{d}(X_i)
\bigr\rangle\leq0 \bigr\}.
\]
We want to show that
minimization of $\hat{L}_{n}(w; \zeta_{d_n})$ over the class
$\mathcal{C}^{(d_n)}$ for increasing $(d_n)$ leads to a
universally consistent classifier for our latent position graphs
setting. However, the loss function
$L(f) = \mathbb{P}(\operatorname{sign}(f(X)) \neq Y)$ of a
classifier $f$ as well as its
empirical version $\hat{L}_n(f)$ is based on the 0--1 loss, which is
discontinuous at $f = 0$. Furthermore, the distribution of
$\zeta_{d}(X)$ not available. This induces complications in relating
$\hat{L}_n(w; \zeta_{d})$ to $\hat{L}_n(w; \Phi_{d})$. That is,
the classifier obtained by minimizing the 0--1 loss using
$\zeta$ might be very different from the classifier
obtained by minimizing the 0--1 loss using~$\Phi$.

To circumvent this issue, we will work with some convex loss function
$\varphi$ that is a surrogate of the 0--1 loss. The notion of
constructing classification algorithms that correspond to minimization
of a convex surrogate for the 0--1 loss is a powerful one and the
authors of
\cite{bartlett06convex,zhang04statis,lugosi04b}, among others,
showed that one can obtain, under appropriate regularity conditions,
Bayes-risk consistent classifiers in this manner.

Let $\varphi\dvtx\mathbb{R} \mapsto[0, \infty)$. We define
the $\varphi$-risk of $f \dvtx\mathcal{X} \mapsto\mathbb{R}$ by
\[
R_{\varphi}(f) = \mathbb{E} \varphi\bigl(Yf(X)\bigr).
\]
Given some data $\mathcal{D}_n =
\{(X_i,Y_i)\}_{i=1}^{n}$, the empirical $\varphi$-risk of $f$ is
defined as
\[
\hat{R}_{\varphi,n}(f) = \frac{1}{n} \sum_{i=1}^{n}
\varphi\bigl(Y_i f(X_i)\bigr).
\]
We will often write $\hat{R}_{\varphi}(f)$ if the number of samples
$(X_i,Y_i)$ in $\mathcal{D}_n$ is clear from the context. Let $w \in
\mathbb{R}^{d}, \|w \| \leq d$ index a linear classifier on
$\mathcal{C}^{(d)}$. Denote by $R_{\varphi}(w; \Phi_{d})$,
$\hat{R}_{\varphi,n}(w;\Phi_{d})$,
$\hat{R}_{\varphi,n}(w;\zeta_{d})$ and
$R_{\varphi,n}(w;\zeta_{d})$ the various quantities analogous to
$L(w; \Phi_{d})$, $\hat{L}_n(w; \Phi_{d})$, $\hat{L}_{n}(w;
\zeta_{d})$ and $L_n(w; \zeta_{d})$ for 0--1 loss defined
previously. Let us also define $R_{\varphi}^{*}$ as the minimum
$\varphi$-risk over all measurable functions $f \dvtx\mathcal{X}
\mapsto\mathbb{R}$.

In this paper, we will assume that the convex surrogate $\varphi
\dvtx\mathbb{R} \mapsto[0,\infty)$ is differentiable with
$\varphi'(0) < 0$. This implies that $\varphi$ is \emph
{classification-calibrated}~\cite{bartlett06convex}. Examples of
classification-calibrated loss functions are the exponential
loss function $\varphi(x) = \exp(-x)$ in boosting, the logit function
$\varphi(x) = \log_{2}(1 + \exp(-x))$ in logistic regression and the
square error loss $\varphi(x) = (1 - x)^2$. For
classification-calibrated loss functions, we have the following
result.
%
%
\begin{theorem}[(\cite{bartlett06convex})]
\label{thm7}
Let $\varphi\dvtx\mathbb{R} \mapsto[0,\infty)$ be
classification-calibrated. 
Then for any sequence of measurable functions $f_i \dvtx
\mathcal{X} \mapsto\mathbb{R}$ and every probability distribution
$F_{\mathcal{X},\mathcal{Y}}$, $R_{\varphi}(f_i) \rightarrow
R_{\varphi}^{*}$ implies $L(f_i) \rightarrow L^{*}$.
\end{theorem}
We now state the main result of this section, which is that
empirical $\varphi$-risk minimization over the class
$\mathcal{C}^{(d_n)}$ for some diverging sequence $(d_n)$ yields a universally
consistent classifier for the latent position graphs setting.
%
%
\begin{theorem}
\label{thm6}
Let $\varepsilon\in(0,1/4)$ be fixed. For a given $d$, let $C_d =\break
\max\{ \varphi'(-d), \varphi'(d)\}$. Suppose that $d_n$ is given by
the following rule:
%
%
\begin{equation}
\label{eq69}\quad d_n = \max\biggl\{d \leq n \dvtx\frac{1}{n}
\bigl(\lambda_{d}(\mathbf{A}) - \lambda_{d+1}(\mathbf{A})
\bigr) \geq32 \sqrt{d C_d} \biggl(\frac{d
\log{n}}{n}
\biggr)^{1/4 - \varepsilon} \biggr\}.
\end{equation}
Let $\tilde{g}_{n}$ be the classifier obtained by empirical
$\varphi$-risk minimization over $\mathcal{C}^{(d_n)}$.
Then $R_{\varphi,n}(\tilde{g}_n) \rightarrow R_{\varphi}^{*}$ as $n
\rightarrow\infty$ and $\operatorname{sign}(\tilde{g}_{n})$ is
universally consistent, that is,
\[
\mathbb{E}\bigl[ \mathbb{P}\bigl(\operatorname{sign}\bigl(\tilde{g}_{n}
\bigl(\zeta_{d_n}(X)\bigr)\bigr) \neq Y | \mathcal{D}_n
\bigr)\bigr] \rightarrow L^{*}
\]
as $n \rightarrow\infty$ for any distribution $F_{\mathcal{X},Y}$.
\end{theorem}

\begin{remark*}
We note that due to the use of the estimated $\zeta$ in place of the
true~$\Phi$, Theorem~\ref{thm6} is limited in two key aspects. The
first is that we do not claim that $\tilde{g}_{n}$ is universally
\emph{strongly} consistent for any $F_{\mathcal{X},Y}$ and the second is
that we cannot specify $d_n$ in advance. In return, the minimization
of the empirical $\varphi$-risk over the class $\mathcal{C}^{(d)}$
is a convex optimization problem and the solution can be obtained
more readily than the minimization of empirical 0--1 loss. For
example, by using squared error loss instead of 0--1 loss, the
classifier that minimizes the empirical $\varphi$-risk can be viewed
as a ridge regression problem. We note also that as the only
accumulation point in the spectrum of $\mathscr{K}$ is at zero, the
sequence $(d_n)$ as specified in equation (\ref{eq69})
exists. Furthermore, such a sequence is only one possibility among
many. In particular, the conclusion in Theorem~\ref{thm6} holds for
any sequence $(d_n)$ that diverges and satisfies the
condition $\delta_{d_n}^{2} = o( n^{-1/2} d^{3/2}
\sqrt{\log{n}})$. Choosing the right $(d_n)$ requires balancing the
approximation error $\inf_{g \in\mathcal{C}^{(d_n)}} R_{\varphi}(g)
- R_{\varphi}^{*}$ and the estimation error
$R_{\varphi}(\tilde{g}_n) - \inf_{w \in\mathcal{C}^{(d_n)}}
R_{\varphi}(g)$, and this can be done using an approach based on
structural risk minimization; see, for example, Section~18.1 of
\cite{devroye1996probabilistic} and~\cite{lugosi04b}.
\end{remark*}
We now proceed to prove Theorem~\ref{thm6}. A rough sketch of the
argument goes as follows. First we show that any classifier $g$ using
the estimated vectors $\zeta_{d}$ induces a classifier $g'$ using the
true truncated feature map $\Phi_{d}$ such that the empirical
$\varphi$-risk of $g$ is ``close'' to the empirical $\varphi$-risk of
$g'$. Then by applying a Vapnik--Chervonenkis-type bound for $g'$, we
show that the classifier
$\tilde{g}$ (using $\zeta_d$) selected by empirical $\varphi$-risk
minimization induces a classifier $\hat{g}$ (using $\Phi_d$) with
the $\varphi$-risk of $\hat{g}$ being ``close'' to the minimum
$\varphi$-risk for the classifiers in the class
$\mathcal{C}^{(d)}$. Universal consistency of $\hat{g}$ and hence of
$\tilde{g}$ follows by letting $d$ grow in a specified manner.

Let $1 \leq d \leq n$. Let $\mathbf{U}_{\mathbf{A}}
\mathbf{S}_{\mathbf{A}}^{1/2}$ be the embedding of $\mathbf{A}$ into
$\mathbb{R}^{d}$. Let $\mathbf{W}_{d} \in\mathcal{M}_{d}(\mathbb{R})$
be an orthogonal matrix given by
\[
\mathbf{W}_{d} = \min_{\mathbf{W} \dvtx\mathbf{W}^{T} \mathbf{W}
= \mathbf{I}} \bigl\| \mathbf{U}_{\mathbf{A}}
\mathbf{S}_{\mathbf{A}}^{1/2} \mathbf{W} - \bolds{\Phi}_{d}
\bigr\|_{F}.
\]

The following result states that if there is a gap in the spectrum of
$\mathscr{K}$ at $\lambda_{d}(\mathscr{K})$, then
$\hat{R}_{\varphi,n}(w; \zeta_{d})$ and
$\hat{R}_{\varphi,n}(\mathbf{W}_{d} w, \Phi_{d})$ is close for all
$w \in\mathbb{R}^{d}, \|w \| \leq d$. That is, the empirical $\varphi
$-risk of a
linear classifier using $\zeta_d$ is not too different from the
empirical $\varphi$-risk of a related classifier (the relationship is
given by~$\mathbf{W}_d$) using $\Phi_d$.
%
%
\begin{proposition}
\label{prop9}
Let $d \geq1$ be such that $\lambda_{d}(\mathscr{K}) >
\lambda_{d+1}(\mathscr{K})$, and let $C_d = \max\{\varphi'(d),
\varphi'(-d)\}$. Then for any $w \in\mathbb{R}^{d}$,
$\| w \| \leq d$, we have, with probability at least $1 - 1/n^{2}$,
%
%
\begin{equation}
\label{eq49} \bigl| \hat{R}_{\varphi, n}(w; \zeta_{d}) -
\hat{R}_{\varphi,n}(\mathbf{W}_{d} w; \Phi_{d})\bigr| \leq27
\delta_{d}^{-2} d C_d \sqrt{\frac{3 d \log{n}}{n}}.
\end{equation}
\end{proposition}
\begin{pf}
We have
\begin{eqnarray*}
&&
\hat{R}_{\varphi, n}(w; \zeta_{d}) - \hat{R}_{\varphi,n}\bigl(
\mathbf{W}^{d} w; \Phi_{d}\bigr) \\
&&\qquad= \frac{1}{n} \sum
_{i=1}^{n} \varphi\bigl(Y_i
\bigl\langle w, \zeta_{d}(X_i)\bigr\rangle\bigr) - \varphi
\bigl(Y_i \bigl\langle\mathbf{W}^{d} w,
\Phi_{d}(X_i) \bigr\rangle\bigr).
\end{eqnarray*}
Now $\varphi$ is convex and thus locally
Lipschitz-continuous. Also, $| \langle w, \Phi_{d}(X) \rangle
| \leq d$ for all $X \in\mathcal{X}$. Hence, there exists a
constant $M$ independent of $n$ and $F_{\mathcal{X},\mathcal{Y}}$
such that
\begin{eqnarray*}
&&\bigl| \varphi\bigl(Y_i \bigl\langle w, \zeta_{d}(X_i)
\bigr\rangle\bigr) - \varphi\bigl(Y_i \bigl\langle
\mathbf{W}_{d} w, \Phi_{d}(X_i) \bigr\rangle
\bigr)\bigr| \\
&&\qquad\leq M \biggl| Y_i \biggl\langle\frac{w}{\|w\|},
\zeta_{d}(X_i)\biggr\rangle- Y_i \biggl\langle
\mathbf{W}^{(d)} \frac{w}{\|w\|}, \Phi_{d}(X_i)
\biggr\rangle\biggr|
\end{eqnarray*}
for all $i$. Thus, by Theorem~\ref{thm1}, we have
%
%
\begin{eqnarray}
\label{eq52}
&&
\bigl|\hat{R}_{\varphi, n}(w; \zeta_{d}) -
\hat{R}_{\varphi,n}(\mathbf{W}_{d} w; \Phi_{d})\bigr|\nonumber\\
&&\qquad\leq
\frac{M}{n} \sum_{i=1}^{n} \biggl|
Y_i \biggl\langle\frac{w}{\|w\|}, \zeta_{d}(X_i)
\biggr\rangle- Y_i \biggl\langle\mathbf{W}_{d}
\frac{w}{\|w\|}, \Phi_{d}(X_i)\biggr\rangle\biggr|
\nonumber
\\
&&\qquad\leq \frac{M}{n} \sum
_{i=1}^{n} \bigl\| \zeta_{d}(X_i) -
(\mathbf{W}_{d})^{T} \Phi_{d}(X_i) \bigr\|
\nonumber\\[-8pt]\\[-8pt]
&&\qquad\leq \frac{M}{\sqrt{n}} \Biggl( \sum_{i=1}^{n}
\bigl\| \zeta_{d}(X_i) - (\mathbf{W}_{d})^{T}
\Phi_{d}(X_i) \bigr\|^{2} \Biggr)^{1/2}
\nonumber
\\
&&\qquad\leq \frac{M}{\sqrt{n}} \bigl\| \mathbf{U}_{\mathbf{A}} \mathbf
{S}_{\mathbf{A}}^{1/2} - (\mathbf{W}_{d})^{T} \bolds{
\Phi}_{d} \bigr\|_{F}
\nonumber
\\
&&\qquad\leq 27 \delta_{d}^{-2} M \sqrt{\frac{3 d \log{n}}{n}}
\nonumber
\end{eqnarray}
with probability at least $1 - 1/n^2$. By the mean-value
theorem, we can take $M = d \max\{\varphi'(d),
\varphi'(-d)\}$ to complete the proof.
\end{pf}

The Vapnik--Chervonenkis theory for 0--1 loss function can also be
extended to the convex surrogate setting $R_{\varphi}$. In particular,
the following result provides a uniform deviation bound for
$| R_{\varphi}(f) - \hat{R}_{\varphi,n}(f)|$ for functions $f$ in some
class $\mathscr{F}$ in terms of the VC-dimension of $\mathscr{F}$.
%
%
\begin{lemma}[(\cite{lugosi04b})]
\label{lem1}
Let $\mathscr{F}$ be a class of functions with VC-dimension $V <
\infty$. Suppose that the range of any $f \in\mathcal{F}$ is contained
in the interval $[-d,d]$. Let $n \geq5$. Then we
have, with probability at least $1 - 1/n^2$,
%
%
\begin{equation}
\label{eq53} \sup_{f \in\mathcal{F}} \bigl| R_{\varphi}(f) -
\hat{R}_{\varphi,n}(f)\bigr| \leq10 d \max\bigl\{\varphi'(d),
\varphi'(-d)\bigr\} \sqrt{\frac{3 V \log{n}}{n}}.
\end{equation}
\end{lemma}

The following result combines Proposition~\ref{prop9} and
Lemma~\ref{lem1} and shows that minimizing
$\hat{R}_{\varphi,n}(w;\zeta_{d})$ over $w \in\mathbb{R}^{d}, \| w
\|
\leq d$ leads
to a classifier whose $\varphi$-risk is close to optimal in the class
$\mathcal{C}^{(d)}$ with high probability.
%
%
\begin{lemma}
\label{lem2}
Let $d \geq1$ be such that $\lambda_{d}(\mathscr{K}) >
\lambda_{d+1}(\mathscr{K})$ and let $C_d = \max\{\varphi'(d),
\varphi'(-d)\}$. Let $\tilde{w}_{d}$ minimize
$\hat{R}_{\varphi,n}(w;\zeta_{d})$ over $\mathbb{R}^{d}, \|w \|
\leq
d$. Then with probability at least $ 1 - 2/n^2$,
%
%
\begin{equation}
\label{eq54} R_{\varphi}(\mathbf{W}_{d} \tilde{w}_{d};
\Phi_{d}) - \inf_{w \in\mathcal{C}^{(d)}}R_{\varphi}( w;
\Phi_{d}) \leq74 \delta_d^{-2} d C_d
\sqrt{\frac{3 d \log{n}}{n}}.
\end{equation}
\end{lemma}
\begin{pf}
For ease of notation, we let $\varepsilon(n,d)$ be the term in the
right-hand side of equation (\ref{eq49}), and let $C(n,d)$ be the
term in
the right-hand side of equation (\ref{eq53}). Also let $\bar
{w}^{(d)}:=
\arginf_{w \in\mathcal{C}^{(d)}} R_{\varphi}(w; \Phi_{d})$. We
then have
\begin{eqnarray*}
R_{\varphi}(\mathbf{W}_{d} \tilde{w}_{d};
\Phi_{d}) &\leq& \hat{R}_{\varphi,n}(\mathbf{W}_{d}
\tilde{w}_{d}; \Phi_{d}) + C(n,d)
\\
&\leq& \hat{R}_{\varphi,n}(\tilde{w}_{d}; \zeta_{d}) +
\varepsilon(n,d) + C(n,d)
\\
&\leq& \hat{R}_{\varphi,n}\bigl((\mathbf{W}_{d})^{T}
\bar{w}_{d}; \zeta_{d}\bigr) + \varepsilon(n,d) + C(n,d)
\\
&\leq& \hat{R}_{\varphi,n}( \bar{w}_{d}; \Phi_{d}) + 2
\varepsilon(n,d) + C(n,d)
\\
&\leq& R_{\varphi}(\bar{w}_{d}; \Phi_{d}) + 2
\varepsilon(n,d) + 2 C(n,d)
\end{eqnarray*}
with probability at least $1 - 2/n^2$.
\end{pf}
\begin{remark*}
Equation (\ref{eq54}) is a VC-type bound. The term $d^{3/2}
\delta_{d}^{-2}$ in equation (\ref{eq54}) can be viewed as contributing
to the generalization error for the classifiers in
$\mathcal{C}^{(d)}$. That is, because we are training using the
estimated vectors in $\mathbb{R}^{d}$, the generalization error not
only depends on the dimension of the embedded space, but also
depends on how accurate the estimated vectors are in that space.
\end{remark*}

We now have the necessary ingredients to prove the main result of this
section.
\begin{pf*}{Proof of Theorem~\ref{thm6}}
Let $(d_n)$ be a nondecreasing sequence of positive integers that
diverges to $\infty$ and that
%
%
\begin{equation}
\label{eq70} \delta_{d_n}^{-2} d_n
C_{d_n} \sqrt{ \frac{d \log{n}}{n}} = o(1).
\end{equation}
By Lemma~\ref{lem2} and the Borel--Cantelli lemma, we have
\[
\lim_{n \rightarrow\infty} \Bigl[ R_{\varphi}(\mathbf{W}_{d_n}
\tilde{w}_{d_n}; \Phi_{d_n}) - \inf_{w \in\mathcal{C}^{(d_n)}}
R_{\varphi}(w; \Phi_{d_n}) \Bigr] = 0
\]
almost surely. As $(d_n)$ diverges, $\lim_{n
\rightarrow\infty} \inf_{w \in\mathcal{C}^{(d_n)}} R_{\varphi}(w;
\Phi_{d_n}) =
R_{\varphi}^{*}$ by Proposition~\ref{prop10}. We therefore have
\[
\lim_{n \rightarrow\infty} R_{\varphi}(\mathbf{W}_{d_n}
\tilde{w}_{d_n}; \Phi_{d_n}) = R_{\varphi}^{*}
\]
almost surely. Now fix a $n$. The empirical $\varphi$-risk
minimization on $w \in\mathbb{R}^{d_n}, \|w \| \leq d_n$ using the
estimated vectors $\zeta_{d_n}$ gives us a classifier $\langle
\tilde{w}_{d_n}, \zeta_{d_n} \rangle$. We now consider the
difference $R_{\varphi,n}(\tilde{w}_{d_n}; \zeta_{d_n}) -
R_{\varphi}(\mathbf{W}_{d_n} \tilde{w}_{d_n}; \Phi_{d_n})$. By a
similar computation to that used in the derivation of
equation (\ref{eq52}), we have
\begin{eqnarray*}
&&
R_{\varphi,n}(\tilde{w}_{d_n}; \zeta_{d_n}) -
R_{\varphi}(\mathbf{W}_{d_n} \tilde{w}_{d_n};
\Phi_{d_n}) \\
&&\qquad= \bigl|\mathbb{E}\bigl[ \varphi\bigl(Y \bigl\langle
\tilde{w}_{d_n}, \zeta_{d_n}(X) \bigr\rangle\bigr)\bigr] -
\mathbb{E}\bigl[ \varphi\bigl(Y \bigl\langle\mathbf{W}_{d_n}
\tilde{w}_{d_n}, \Phi_{d_n}(X)\bigr\rangle\bigr)\bigr]\bigr|
\\
&&\qquad\leq d_{n} C_{d_n} \mathbb{E}\bigl[ \bigl\|
\zeta_{d_n}(X) - (\mathbf{W}_{d_n})^{T}
\Phi_{d_n}(X) \bigr\|\bigr]
\\
&&\qquad\leq d_{n} C_{d_n} \sqrt{ \mathbb{E}\bigl[ \bigl\|
\zeta_{d_n}(X) - (\mathbf{W}_{d_n})^{T}
\Phi_{d_n}(X) \bigr\|^{2}\bigr]}
\\
&&\qquad\leq 27 \delta_{d_n}^{-2} d_{n} C_{d_n}
\sqrt{ \frac{ 6 d
\log{n}}{n}} = o(1).
\end{eqnarray*}
We therefore have
\[
\lim_{n \rightarrow\infty} R_{\varphi,n}(\tilde{w}_{d_n};
\zeta_{d_n}) = R_{\varphi}^{*}.
\]
Thus, by Theorem~\ref{thm7}, we have
\[
\lim_{n \rightarrow\infty} \mathbb{E}\bigl[L_{n}(
\tilde{w}_{d_n}; \zeta_{d_n})\bigr] = L^{*}.
\]
The only thing that remains is the use of
$\frac{1}{n}(\lambda_{d}(\mathbf{A}) -
\lambda_{d+1}(\mathbf{A}))$ as an estimate for~$\delta_{d}$. By
Proposition~\ref{prop3} and Theorem~\ref{thm1}, we have
%
%
\begin{equation}
\label{eq78} \sup_{d \geq1} \biggl|\delta_{d} -
\frac{1}{n}\bigl(\lambda_{d}(\mathbf{A}) - \lambda_{d+1}(
\mathbf{A})\bigr)\biggr| \leq10 \sqrt{\frac{\log{(2/n^2)}}{n}}
\end{equation}
with probability at least $1 - 2/n^2$. Thus, if $d_n$ satisfy
equation (\ref{eq69}), then equation (\ref{eq78}) implies that
equation (\ref{eq70}) holds for $d_n \rightarrow\infty$
with probability at least $1 - 2/n^2$.
Finally, we note that as $n \rightarrow
\infty$, there exists a sequence $(d_n)$ that satisfies
equation (\ref{eq69}) and diverges
to $\infty$, as the only accumulation point in the spectrum of
$\mathscr{K}$ is at zero.
\end{pf*}
%
\section{Conclusions}
\label{secconclusion}
In this paper we investigated the problem of finding a universally
consistent classifier for classifying the vertices of latent position
graphs. We showed that if the link function $\kappa$ used in the
construction of the graphs belong to the class of universal kernels,
then an empirical $\varphi$-risk minimization approach, that is,
minimizing a
convex surrogate of the 0--1 loss over the class of linear
classifiers in $\mathbb{R}^{d_n}$ for some sequence $d_n \rightarrow
\infty$, yields universally consistent vertices classifiers.

We have presented the universally consistent classifiers in the
setting where the graphs are on $n+1$ vertices, there are $n$ labeled
vertices and the task is to classify the remaining unlabeled
vertex. It is easy to see that in the case where there are only $m <
n$ labeled vertices, the same procedure given in
Theorem~\ref{thm6} with $n$ replaced by $m$ still yields universally
consistent classifiers, provided that $m \rightarrow\infty$.

The bound for the generalization error of the classifiers in
Section~\ref{seccons-vert-class} is of the form $O(n^{-1/2}
\delta_{d}^{-2} \sqrt{d^{3} \log{n}})$. This
bound depends on both the subspace projection error in
Section~\ref{secestim-feat-maps} as well as the generalization error of
the class $\mathcal{C}^{(d_n)}$. It is often the case that the bound on
the generalization error of the class $\mathcal{C}^{(d_n)}$ can be
improved, as long as the classification problems satisfy a
``low-noise'' condition, that is, that the posterior probability $\eta(x)
= \mathbb{P}[Y = 1 | X = x]$ is bounded away from $1/2$. Results on
fast convergence rates in low-noise conditions, for example,
\cite{blanchard08statisperfor,bartlett06convex}
can thus be used, but as the subspace
projection error is independent of the low-noise condition, there
might not be much improvement in the resulting error bound.

Also related to the above issue is the choice of the sequence
$(d_n)$. If more is known about the kernel $\kappa$, then the choice
for the sequence $(d_n)$ can be adjusted accordingly. For example, good
bounds for $\Lambda_{k} = \sum_{j \geq k} \lambda_{j}(\mathscr{K})$,
the sum of the tail eigenvalues of $\mathscr{K}$, along with bounds
for the error between the truncated feature map $\Phi_{d}$ and the
feature map $\Phi$ from
\cite
{braun06accurerrorboundeigenkernelmatrix,shawe-taylor05eigengramgenerpca,blanchard07statis}
can be used to select the sequence~$(d_n)$.

The results presented in Section~\ref{secestim-feat-maps} and
Section~\ref{seccons-vert-class} implicitly assumed that the graphs
arising from the latent position model are dense. It is possible to
extend these results to sparse graphs. A sketch of the ideas is as
follows. Let $\rho_n \in(0,1)$ be a scaling parameter, and consider
the latent position model with kernel $\kappa$ and distribution $F$
for the latent features $\{X_i\}$. Given $\{X_i\}_{i=1}^{n}
\stackrel{\mathrm{i.i.d.}}{\sim} F$, let $\mathbf{K}_n = (\rho_n
\kappa(X_i,X_j))_{i,j=1}^{n}$, that is, the entries of $\mathbf{K}_n$ are
given by the kernel $\kappa$ scaled by the scaling parameter
$\rho_n$. This variant of the latent position model is also present in
the notion of inhomogeneous random graphs
\cite{bollobas07,oliveira2010concentration}. Given $\mathbf{K}_n$,
$\mathbf{A}_n = \operatorname{Bernoulli}(\mathbf{K}_n)$ is the adjacency
matrix. The factor $\rho_n$ controls the sparsity of the resulting
latent position graph. For example, $\rho_n = (\log{n})/n$ leads to
sparse, connected graphs almost surely while $\rho_n = 1/n$ leads to
graphs with a single giant connected component
\cite{bollobas07}. Suppose now that $\rho_n = \Omega((\log{n})/n)$. The
following result is a restatement of Theorem~\ref{thm1} for the
latent position model in the presence of the scaling parameter
$\rho_n$. Its proof is almost identical to that of Theorem~\ref{thm1}
provided that one uses the bound in term of the maximum degree
$\Delta$ in Proposition~\ref{prop3}. We note that $\delta_{d}$ is
defined in terms of the spectrum of $\mathscr{K}$ which does not
depend on the scaling parameter $\rho_n$, and similarly for the
feature map $\Phi$ and its truncation $\Phi_{d}$.
%
%
\begin{theorem}
\label{thm9}
Let $d \geq1$ be given. Denote by $\delta_{d}$ the quantity
$\lambda_{d}(\mathscr{K}) - \lambda_{d+1}(\mathscr{K})$, and suppose
that $\delta_{d} > 0$. Then with probability greater than $1-2
\eta$, there exists a unitary matrix $\mathbf{W} \in
\mathcal{M}_{d}(\mathbb{R})$ such that
%
%
\begin{equation}\label{eq1}
\bigl\| \rho_n^{-1/2} \mathbf{U}_\mathbf{A}
\mathbf{S}_\mathbf{A}^{1/2} \mathbf{W} - \bolds{\Phi}_{d}
\bigr\|_{F} \leq27 \delta_{d}^{-2} \sqrt{
\frac{d
\log{(n/\eta)}}{\rho_n}},
\end{equation}
where $\bolds{\Phi}_{d}$ denotes the matrix in
$\mathcal{M}_{n,d}(\mathbb{R})$ whose $i$th row is
$\Phi_{d}(X_i)$. Let us denote by $\hat{\Phi}_{d}(X_i)$ the $i$th row
of $\rho_n^{-1/2}
\mathbf{U}_\mathbf{A}\mathbf{S}_\mathbf{A}^{1/2}\mathbf{W}$. Then, for
each $i\in[n]$ and any $\varepsilon> 0$,
%
%
\begin{equation}
\label{eq2} \Pr\bigl[\bigl\|\hat{\Phi}_{d}(X_i) -
\Phi_{d}(X_i) \bigr\| > \varepsilon\bigr] \leq27
\delta_d^{-2} \varepsilon^{-1} \sqrt{
\frac{6d \log{n}}{n \rho_n}}.
\end{equation}
\end{theorem}
Thus, for $\rho_n = n^{-1+\varepsilon}$ for some $\varepsilon
> 0$ [or even $\rho_n = (\log^{k}{n})/n$ for some sufficient large
$k$], equation (\ref{eq2}) states that with high probability, the estimated
feature map is (after scaling by $\rho_n^{-1/2}$ and rotation)
converging to the true truncated feature map $\Phi_d$ as $n
\rightarrow\infty$. The results from Section~\ref
{seccons-vert-class} can
then be modified to show the existence of a universally consistent
linear classifier. The main difference between the sparse setting and
the dense setting would be the generalization bounds in
Proposition~\ref{prop9} and Lemma~\ref{lem2}. This would lead to a
different selection rule for the sequence of embedding dimensions $d_n
\rightarrow\infty$ then the one in Theorem~\ref{thm9}, that is, the
$d_n$ would diverge more slowly for the sparse setting compared to the
dense setting. A~precise statement and formulation of the results in
Section~\ref{seccons-vert-class} for the sparse setting might require some
care, but should be for the most part straightforward. We also note
that even though $\rho_n$ is most likely unknown, one can scale the
embedding $\mathbf{Z} = \mathbf{U}_{\mathbf{A}}
\mathbf{S}_{\mathbf{A}}^{1/2}$ by any value $c_n$ that is of the same
order as $\rho_n^{-1/2}$. An appropriate value for $c_n$ is, for
example, one that makes $\max_{i} \| c_n Z_i \|^{2} = 1$ where $Z_i$
is the $i$th row of $\mathbf{Z}$.

Finally, we note it is of potential interest to extend the results
herein to graphs with attributes on the edges, latent position graphs
with nonpositive definite link functions $\kappa$ and graphs with
errorfully observed edges.

%
%
\begin{appendix}\label{app}
\section{Additional proofs}
\label{secadditional-proofs}
\begin{pf*}{Proof of Lemma~\ref{lem4}}
Let $\Psi_{r,n} \in\mathbb{R}^{n}$ be the vector whose entries
are $\sqrt{\lambda}_{r}\psi_{r}(X_i)$ for $i = 1,2,\ldots,n$ with
$\lambda_{r} = \lambda_{r}(\mathscr{K})$. We note that $\mathbf{K} =
\sum_{r=1}^{\infty} \Psi_{r,n} \*\Psi_{r,n}^{T}$. Let $\hat
{u}^{(1)},\ldots, \hat{u}^{(d)}$ be the eigenvectors associated
with the $d$ largest eigenvalues of $\mathbf{K}/n$. We have
\[
\mathcal{P}_{\mathbf{K}} \mathbf{K} = \sum_{s=1}^{d}
\sum_{r=1}^{\infty} \hat{u}^{(s)}
\bigl(\hat{u}^{(s)}\bigr)^{T} \Psi_{r,n}
\Psi_{r,n}^{T} \hat{u}^{(s)} \bigl(
\hat{u}^{(s)}\bigr)^{T}.
\]
The $ij$th entry of $\mathcal{P}_{\mathbf{K}} \mathbf{K}$ is then given
by
\[
\sum_{s=1}^{d} \sum
_{r=1}^{\infty} \hat{u}^{(s)}_{i}
\bigl(\hat{u}^{(s)}\bigr)^{T} \Psi_{r,n}
\Psi_{r,n}^{T} \hat{u}^{(s)} \hat{u}^{(s)}_{j}.
\]
Let $\hat{v}^{(1)},\ldots, \hat{v}^{(d)}$ be the extensions of
$\hat{u}^{(1)},\ldots, \hat{u}^{(d)}$ as defined by equation (\ref{eq36}).
We then have, for any $s =1,2,\ldots,d$,
\begin{eqnarray*}
\bigl\langle\hat{v}^{(s)}, \sqrt{\lambda}_{r}
\psi_{r} \bigr\rangle_{\mathscr{H}} &=& \Biggl\langle
\frac{1}{\sqrt{\hat{\lambda}_{s}
n}} \sum_{i=1}^{n} \kappa(
\cdot, X_i) \hat{u}^{(s)}_i, \sqrt{
\lambda}_{r} \psi_{r} \Biggr\rangle_{\mathscr{H}}
\\
&=& \Biggl\langle\frac{1}{\sqrt{\hat{\lambda}_{s} n}} \sum_{i=1}^{n}
\sum_{r'} \sqrt{\lambda}_{r'}
\psi_{r'}(X_i) \sqrt{\lambda}_{r'}
\psi_{r'} \hat{u}^{(s)}_i, \sqrt{
\lambda}_{r} \psi_{r} \Biggr\rangle_{\mathscr{H}}
\\
&=& \frac{1}{\sqrt{\hat{\lambda}_{s}n}} \sum_{i=1}^{n}
\psi_{r}(X_i) \sqrt{\lambda}_{r}
\hat{u}^{(s)}_i \\
&=& \frac{1}{\sqrt{\hat{\lambda}_{s}n}} \bigl\langle
\hat{u}^{(s)}, \Psi_{r,n} \bigr\rangle_{\mathbb{R}^{n}}.
\end{eqnarray*}
We thus have
%
%
\begin{equation}
\label{eq39} \hat{u}^{(s)}_i \bigl(\hat{u}^{(s)}
\bigr)^{T} \Psi_{r,n} = \hat{u}^{(s)}_i
\bigl\langle\hat{u}^{(s)}, \Psi_{r,n} \bigr
\rangle_{\mathbb{R}^{n}} = \hat{v}^{(s)}(X_i) \bigl\langle
\hat{v}^{(s)}, \sqrt{\lambda_r} \psi_{r} \bigr
\rangle_{\mathscr{H}}.
\end{equation}
Now let $\xi^{(s)}(X) = \sum_{r = 1}^{\infty} \langle\hat{v}^{(s)},
\psi_{r} \sqrt{\lambda}_{r} \rangle_{\mathscr{H}} \hat{v}^{(s)}(X)
\sqrt{\lambda}_{r} \psi_{r} \in\mathscr{H}$. $\xi^{(s)}(X)$ is
the embedding of the sequence $(\langle\hat{v}^{(s)},
\sqrt{\lambda}_{r} \psi_{r} \rangle_{\mathscr{H}}
\hat{v}^{(s)}(X))_{r=1}^{\infty} \in l_2$ into $\mathscr{H}$;
see equation (\ref{eq82}). By
equation (\ref{eq39}) and the definition of $\langle\cdot, \cdot
\rangle_{\mathscr{H}}$ [equation (\ref{eq83})], the
$ij$th entry of $\mathcal{P}_{\mathbf{K}} \mathbf{K}$ can be
written as
\[
\sum_{s=1}^{d} \sum
_{r=1}^{\infty} \hat{u}^{(s)}_{i}
\bigl(\hat{u}^{(s)}\bigr)^{T} \Psi_{r,n}
\Psi_{r,n}^{T} \hat{u}^{(s)} \hat{u}^{(s)}_{j}
= \sum_{s=1}^{d} \bigl\langle
\xi^{(s)}(X_i), \xi^{(s)}(X_j) \bigr
\rangle_{\mathscr{H}}.
\]
We note that, by the
reproducing kernel property of $\kappa(\cdot, x)$,
\begin{eqnarray*}
\xi^{(s)}(X) &=& \sum_{r = 1}^{\infty}
\bigl\langle\hat{v}^{(s)}, \psi_{r} \sqrt{
\lambda}_{r} \bigr\rangle_{\mathscr{H}} \hat{v}^{(s)}(X)
\sqrt{\lambda}_{r} \psi_{r}
\\
&=& \sum_{r = 1}^{\infty} \bigl\langle
\hat{v}^{(s)}, \psi_{r} \sqrt{\lambda}_{r} \bigr
\rangle_{\mathscr{H}} \bigl\langle\hat{v}^{(s)}, \kappa(\cdot, X) \bigr
\rangle_{\mathscr{H}} \sqrt{\lambda}_{r} \psi_{r}
\\
&=& \bigl\langle\hat{v}^{(s)}, \kappa(\cdot, X) \bigr
\rangle_{\mathscr{H}} \sum_{r=1}^{\infty} \bigl
\langle\hat{v}^{(s)}, \psi_{r} \sqrt{\lambda}_{r}
\bigr\rangle_{\mathscr{H}} \sqrt{\lambda}_{r} \psi_{r}
\\
&=& \bigl\langle\hat{v}^{(s)}, \kappa(\cdot, X) \bigr
\rangle_{\mathscr{H}} \hat{v}^{(s)}.
\end{eqnarray*}
As the $\hat{v}^{(s)}$ are orthogonal with respect to $\langle\cdot,
\cdot\rangle_{\mathscr{H}}$, the $ij$th entry of
$\mathcal{P}_{\mathbf{K}} \mathbf{K}$ can also be written as
\begin{eqnarray*}
&&
\sum_{s=1}^{d} \bigl\langle
\xi^{(s)}(X_i), \xi^{(s)}(X_j) \bigr
\rangle_{\mathscr{H}} \\
&&\qquad= \sum_{s=1}^{d}
\bigl\langle\hat{v}^{(s)}, \kappa(\cdot, X_i) \bigr
\rangle_{\mathscr{H}} \bigl\langle\hat{v}^{(s)}, \hat{v}^{(s)}
\bigr\rangle_{\mathscr{H}} \bigl\langle\hat{v}^{(s)}, \kappa(\cdot,
X_j) \bigr\rangle_{\mathscr{H}}
\\
&&\qquad= \sum_{s=1}^{d} \sum
_{s'=1}^{d} \bigl\langle\hat{v}^{(s)},
\kappa(\cdot, X_i) \bigr\rangle_{\mathscr{H}} \bigl\langle
\hat{v}^{(s)}, \hat{v}^{(s')} \bigr\rangle_{\mathscr{H}} \bigl
\langle\hat{v}^{(s')}, \kappa(\cdot, X_j) \bigr
\rangle_{\mathscr{H}}
\\
&&\qquad= \Biggl\langle\sum_{s=1}^{d} \bigl
\langle\hat{v}^{(s)}, \kappa(\cdot, X_i) \bigr
\rangle_{\mathscr{H}} \hat{v}^{(s)}, \sum_{s=1}^{d}
\bigl\langle\hat{v}^{(s)}, \kappa(\cdot, X_j) \bigr
\rangle_{\mathscr{H}} \hat{v}^{(s)} \Biggr\rangle_{\mathscr{H}}
\\
&&\qquad= \bigl\langle\hat{\mathcal{P}}_{d} \kappa(\cdot, X_i),
\hat{\mathcal{P}}_{d} \kappa(\cdot, X_j) \bigr
\rangle_{\mathscr{H}}.
\end{eqnarray*}
As the $\hat{\mathcal{P}}_{d} \kappa(\cdot, \cdot)$ lies in a
$d$-dimensional
subspace of $\mathscr{H}$, they can be isometrically embedded into
$\mathbb{R}^{d}$. Thus there exists a matrix $\mathbf{X} \in
\mathcal{M}_{n,d}(\mathbb{R})$ such that $\mathbf{X} \mathbf{X}^{T} =
\mathbf{U}_{\mathbf{K}} \mathbf{S}_{\mathbf{K}}
\mathbf{U}_{\mathbf{K}}^{T}$ and that the rows of $\mathbf{X}$
correspond to the projections $\hat{\mathcal{P}}_{d} \kappa(\cdot,
X_i)$. Therefore, there exists a unitary matrix $\mathbf{W} \in
\mathcal{M}_{d}(\mathbf{R})$ such that $\mathbf{X} =
\mathbf{U}_{\mathbf{K}} \mathbf{S}_{\mathbf{K}}^{1/2} \mathbf{W}$
as desired.
\end{pf*}

%
%
\begin{lemma}
\label{lem3}
Let $\mathbf{A}$ and $\mathbf{B}$ be $n \times n$ positive
semidefinite matrices with
$\operatorname{rank}(\mathbf{A}) = \operatorname{rank}(\mathbf{B}) =
d$. Let
$\mathbf{X}, \mathbf{Y} \in\mathcal{M}_{n,d}(\mathbb{R})$ be of
full column rank such that $\mathbf{X} \mathbf{X}^{T} = \mathbf{A}$
and $\mathbf{Y} \mathbf{Y}^{T} = \mathbf{B}$. Let $\delta$ be the
smallest nonzero eigenvalue of~$\mathbf{B}$. Then there exists an
orthogonal matrix $\mathbf{W} \in\mathcal{M}_{d}(\mathbb{R})$ such
that
%
%
\begin{equation}
\label{eq81} \| \mathbf{X} \mathbf{W} - \mathbf{Y} \|_{F} \leq
\frac{ \|
\mathbf{A} - \mathbf{B} \| (\sqrt{d
\| \mathbf{A} \|}+ \sqrt{d \| \mathbf{B} \|)}}{ \delta}.
\end{equation}
\end{lemma}
\begin{pf}
Let $\mathbf{R} = \mathbf{A} - \mathbf{B}$. As $\mathbf{Y}$ is of
full column rank, $\mathbf{Y}^{T} \mathbf{Y}$ is invertible, and its
smallest eigenvalue is $\delta$. We then have
\[
\mathbf{Y} = \mathbf{X} \mathbf{X}^{T} \mathbf{Y} \bigl(
\mathbf{Y}^{T} \mathbf{Y}\bigr)^{-1} - \mathbf{R} \mathbf{Y}
\bigl(\mathbf{Y}^{T} \mathbf{Y}\bigr)^{-1}.
\]
Let $\mathbf{T} = \mathbf{X}^{T} \mathbf{Y} (\mathbf{Y}^{T}
\mathbf{Y})^{-1}$. We then have
\[
\mathbf{T}^{T} \mathbf{T} - \mathbf{I} = \bigl(\mathbf{Y}^{T}
\mathbf{Y}\bigr)^{-1} \mathbf{Y}^{T} \mathbf{X}
\mathbf{X}^{T} \mathbf{Y} \bigl(\mathbf{Y}^{T} \mathbf{Y}
\bigr)^{-1} - \mathbf{I} = \bigl(\mathbf{Y}^{T} \mathbf{Y}
\bigr)^{-1} \mathbf{Y}^{T} \mathbf{R} \mathbf{Y} \bigl(
\mathbf{Y}^{T} \mathbf{Y}\bigr)^{-1}.
\]
Therefore,
\[
- \bigl(\mathbf{Y}^{T} \mathbf{Y}\bigr)^{-1}
\mathbf{Y}^{T} \| \mathbf{R} \| \mathbf{Y} \bigl(\mathbf{Y}^{T}
\mathbf{Y}\bigr)^{-1} \preceq\mathbf{T}^{T} \mathbf{T} -
\mathbf{I} \preceq\bigl(\mathbf{Y}^{T} \mathbf{Y}\bigr)^{-1}
\mathbf{Y}^{T} \| \mathbf{R} \| \mathbf{Y} \bigl(\mathbf{Y}^{T}
\mathbf{Y}\bigr)^{-1},
\]
where $\preceq$ refers to the positive semi-definite ordering for matrices.
We thus have
\[
\bigl\| \mathbf{T}^{T} \mathbf{T} - \mathbf{I} \bigr\|_{F} \leq\|
\mathbf{R}\| \cdot\bigl\| \bigl(\mathbf{Y}^{T} \mathbf{Y}
\bigr)^{-1} \bigr\|_{F} \leq\sqrt{d} \|\mathbf{R} \| \cdot\bigl\|
\bigl(\mathbf{Y}^{T} \mathbf{Y}\bigr)^{-1} \bigr\| \leq
\frac{\| \mathbf{R} \|
\sqrt{d}}{\delta}.
\]
Now let $\mathbf{W}$ be the orthogonal matrix in the polar
decomposition $\mathbf{T} = \mathbf{W}
(\mathbf{T}^{T} \mathbf{T})^{1/2}$. We then have
\begin{eqnarray*}
\| \mathbf{X} \mathbf{W} - \mathbf{Y} \|_{F} &\leq& \| \mathbf{X}
\mathbf{W} - \mathbf{X} \mathbf{T} \|_{F} + \| \mathbf{X} \mathbf{T} -
\mathbf{Y} \|_{F}
\\
&\leq& \| \mathbf{X} \| \cdot\bigl\| \bigl(\mathbf{T}^{T} \mathbf{T}
\bigr)^{1/2} - \mathbf{I} \bigr\|_{F} + \| \mathbf{R} \| \cdot\bigl\|
\mathbf{Y} \bigl( \mathbf{Y}^{T} \mathbf{Y}\bigr)^{-1}
\bigr\|_{F}
\\
&\leq& \| \mathbf{X} \| \cdot\bigl\| \bigl(\mathbf{T}^{T} \mathbf{T}
\bigr)^{1/2} - \mathbf{I} \bigr\|_{F} + \| \mathbf{R} \| \cdot\|
\mathbf{Y} \| \cdot\bigl\|\bigl(\mathbf{Y}^{T} \mathbf{Y}
\bigr)^{-1} \bigr\|_{F}.
\end{eqnarray*}
Now, $\|(\mathbf{T}^{T} \mathbf{T})^{1/2} - \mathbf{I} \|_{F} \leq\|
\mathbf{T}^{T} \mathbf{T} - \mathbf{I} \|_{F}$. Indeed,
\begin{eqnarray*}
\bigl\|\bigl(\mathbf{T}^{T} \mathbf{T}\bigr)^{1/2} - \mathbf{I}
\bigr\|_{F}^{2} &=& \sum_{i=1}^{d}
\bigl(\lambda_{i}\bigl(\mathbf{T}^{T} \mathbf{T}
\bigr)^{1/2} - 1\bigr)^{2}
\leq\sum
_{i=1}^{d} \bigl(\lambda_{i}\bigl(
\mathbf{T}^{T} \mathbf{T}\bigr) - 1\bigr)^{2} \\
&=& \bigl\|
\mathbf{T}^{T} \mathbf{T} - \mathbf{I} \bigr\|_{F}^{2}.
\end{eqnarray*}
We thus have
\[
\| \mathbf{X} \mathbf{W} - \mathbf{Y} \| \leq\bigl(\| \mathbf{X} \| + \|
\mathbf{Y}
\|\bigr) \frac{\| \mathbf{R} \| \sqrt{d}}{\delta},
\]
and equation (\ref{eq81}) follows.
\end{pf}

%
%
\begin{proposition}
\label{prop10}
Let $\kappa$ be a universal kernel on $\mathcal{X}$, and let $\Phi
\dvtx\mathcal{X} \mapsto l_2$ be a feature map of $\kappa$. Let
$\mathcal{C}^{(1)}, \mathcal{C}^{(2)}, \ldots$ be the sequence of
classifiers of the form in equation (\ref{eq11}). Then
%
%
\begin{equation}
\label{eq65} \lim_{d \rightarrow\infty} \inf_{f \in\mathcal{C}^{(d)}}
R_{\varphi}(f) = R_{\varphi}^{*}.
\end{equation}
\end{proposition}
\begin{pf}
We note that this result is a
slight variation of Lemma 1 in~\cite{lugosi04b}. For completeness,
we sketch its proof here. Let $f^{*}$ be the function defined by
\[
f^{*}(x) = \inf_{\alpha\in\mathbb{R}}\bigl\{ \eta(x) \varphi(
\alpha) + \bigl(1 - \eta(x)\bigr)\varphi(- \alpha)\bigr\},
\]
where $\eta(x) = \mathbb{P}[Y = 1 | X = x]$. Then $R_{\varphi}^{*} =
\mathbb{E}[f^{*}]$. Now, for a given $\beta\in[0,1/2]$, let
$H_{\beta} = \{x \dvtx| \eta(x) - 1/2 | > \beta\}$, and let
$\bar{H}_{\beta}$ be the complement of $H_{\beta}$. We consider the
decomposition
\[
R_{\varphi}^{*} = \mathbb{E}\bigl[f^{*}(X) \mathbf{1}\{X
\in H_{\beta}\}\bigr] + \mathbb{E}\bigl[f^{*}(X) \mathbf{1}\{X \in
\bar{H}_{\beta}\}\bigr].
\]
The restriction of $f^{*}$ to $\bar{H}_{\beta}$ is measurable with
range $[-C_{\beta}, C_{\beta}]$ for some finite constant $C_{\beta}
> 0$. The set of functions $\langle w, \Phi\rangle_{\mathscr{H}}$ is
dense in $\mathcal{C}(X)$ and hence also dense in
$L^{1}(\mathcal{X}, F_{\mathcal{X}})$. Thus, for any $\varepsilon> 0$, there
exists a $w \in\mathscr{H}$ such that
\[
\mathbb{E}\bigl[f^{*}(X) \mathbf{1}\{X \in\bar{H}_{\beta}\}\bigr]
- \mathbb{E}\bigl[\bigl\langle w, \Phi(X)\bigr\rangle_{\mathscr{H}}
\mathbf
{1}\{X
\in\bar{H}_{\beta}\}\bigr] < \varepsilon.
\]
Furthermore, $\mathbb{E}[f^{*}(X) \mathbf{1}\{X \in H_{\beta}\}]
\rightarrow0$ as $\beta\rightarrow1/2$. Indeed, $H_{1/2} = \{ x
\dvtx\eta(x) \in\{0,1\}\}$ so we can select $\alpha$ so that
$\varphi(\alpha) = 0$ if $\eta(x) = 1$ and $\varphi(-\alpha) = 0$
if $\eta(x) = 0$. To complete the proof, we note that the
$\mathcal{C}^{(d)}$ are nested, that is, $\mathcal{C}^{(d)} \subset
\mathcal{C}^{(d+1)}$. Hence $\inf_{f \in\mathcal{C}^{(d)}}
R_{\varphi}(f)$ is a decreasing sequence that converges to
$R_{\varphi}^{*}$ as desired.
\end{pf}

\section{Spectra of integral operators and kernel~matrices}
\label{secspectra-mathbfk-spec}
We can tie the spectrum and eigenvectors of $\mathbf{K}$ to the
spectrum and eigenfunctions of $\mathscr{K}$ by constructing an
extension operator $\mathscr{K}_{\mathscr{H},n}$ for $\mathbf{K}$
and relating the
spectra of $\mathscr{K}$ to that of $\mathscr{K}_n$
\cite{rosasco10integoperat}. Let $\mathscr{H}$ be
the reproducing kernel Hilbert space for $\kappa$. Let
$\mathscr{K}_{\mathscr{H}} \dvtx\mathscr{H} \mapsto\mathscr{H}$ and
$\mathscr{K}_{\mathscr{H},n} \dvtx\mathscr{H} \mapsto\mathscr
{H}$ be
the linear operators defined by
\begin{eqnarray*}
\mathscr{K}_{\mathscr{H}} \eta&=& \int_{\mathcal{X}} \bigl\langle\eta,
\kappa(\cdot, x) \bigr\rangle_{\mathscr{H}} \kappa(\cdot, x) \,dF(x),
\\
\mathscr{K}_{\mathscr{H},n} \eta&=& \frac{1}{n} \sum
_{i=1}^{n} \bigl\langle\eta, \kappa(\cdot,
{X_i}) \bigr\rangle_{\mathscr{H}} \kappa(\cdot,
{X_i}).
\end{eqnarray*}
The operators $\mathscr{K}_{\mathscr{H}}$ and
$\mathscr{K}_{\mathscr{H},n}$ are defined on the same Hilbert space
$\mathscr{H}$, in contrast to $\mathscr{K}$ and $\mathbf{K}$ which are
defined on the different spaces $L^{2}(\mathcal{X},F)$ and $\mathbb{R}^{n}$,
respectively. Thus, we can relate the spectra of
$\mathscr{K}_{\mathscr{H}}$ and
$\mathscr{K}_{\mathscr{H},n}$. Furthermore, we can also relate the spectra
of $\mathscr{K}$ and $\mathscr{K}_{\mathscr{H}}$ as well as the
spectra of
$\mathbf{K}$ and $\mathscr{K}_{\mathscr{H},n}$, therefore giving us a
relationship between the spectra of $\mathscr{K}$ and
$\mathbf{K}$. A precise statement of the relationships is contained in the
following results.

%
%
\begin{proposition}[(\cite{rosasco10integoperat,luxburg08consis})]
\label{prop5}
The operators $\mathscr{K}_{\mathscr{H}}$ and
$\mathscr{K}_{\mathscr{H},n}$ are positive, self-adjoint operators
and are of trace class with $\mathscr{K}_{\mathscr{H},n}$ being of
finite rank. The spectra of $\mathscr{K}$ and
$\mathscr{K}_{\mathscr{H}}$ are contained in $[0,1]$ and are the
same, possibly up to the zero eigenvalues. If $\lambda$ is a
nonzero eigenvalue of $\mathscr{K}$ and $u$ and $v$ are associated
eigenfunction of $\mathscr{K}$ and $\mathscr{K}_{\mathscr{H}}$,
normalized to norm $1$ in $L^{2}(\mathcal{X}, F)$ and $\mathscr{H}$,
respectively, then
%
%
\begin{eqnarray}
\label{eq28} u(x) &=& \frac{v(x)}{\sqrt{\lambda}} \qquad\mbox{for $x \in
\operatorname{supp}(F)$};\nonumber\\[-8pt]\\[-8pt]
v(x) &=& \frac{1}{\sqrt{\lambda}} \int_{\mathcal{X}} \kappa
\bigl(x,x'\bigr) u\bigl(x'\bigr) \,dF\bigl(x'\bigr).\nonumber
\end{eqnarray}
Similarly, the spectra of
$\mathbf{K}/n$ and $\mathscr{K}_{\mathscr{H},n}$ are contained in
$[0,1]$ and are the same, possibly up to the zero eigenvalues. If
$\hat{\lambda}$ is a nonzero eigenvalue of $\mathbf{K}$ and
$\hat{u}$ and $\hat{v}$ are the corresponding eigenvector and
eigenfunction of $\mathbf{K}/n$ and
$\mathscr{K}_{\mathscr{H},n}$, normalized to norm $1$ in
$\mathbb{R}^{n}$ and $\mathscr{H}$, respectively, then
%
%
\begin{equation}
\label{eq36} \hat{u}_{i} =
\frac{\hat{v}(x_i)}{\sqrt{\hat{\lambda}}};\qquad
\hat{v}(\cdot) =
\frac{1}{\sqrt{\hat{\lambda}n}} \sum_{i=1}^{n} \kappa(
\cdot,x_i) \hat{u}_{i}.
\end{equation}
\end{proposition}
Equation (\ref{eq36}) in Proposition~\ref{prop5} states that an
eigenvector $\hat{u}$ of $\mathbf{K}/n$, which is only defined for
$X_1, X_2,\ldots, X_n$, can be extended to an eigenfunction $\hat{v} \in
\mathscr{H}$ of $\mathscr{K}_{\mathscr{H},n}$ defined
for all $x \in\mathcal{X}$, and furthermore, that $\hat{u}_i =
\hat{v}(X_i)$ for all $i=1,2,\ldots,n$.
%
%
\begin{theorem}[(\cite{rosasco10integoperat,zwald06})]
\label{thm5}
Let $\tau> 0$ be arbitrary. Then with probability at least $1 -
2e^{-\tau}$,
%
%
\begin{equation}
\label{eq29} \| \mathscr{K}_{\mathscr{H}} - \mathscr{K}_{\mathscr
{H},n}
\|_{\mathrm{HS}} \leq2 \sqrt{2} \sqrt{\frac{\tau}{n}},
\end{equation}
where $\| \cdot\|_{\mathrm{HS}}$ is the Hilbert--Schmidt norm. Let
$\{\lambda_{j}\}$ be a decreasing enumeration of the eigenvalues for
$\mathscr{K}_{\mathscr{H}}$, and let $\{\hat{\lambda}_j\}$ be an
extended decreasing enumeration of $\mathscr{K}_{\mathscr{H},n}$;
that is, $\hat{\lambda}_j$ is either an eigenvalue of
$\mathscr{K}_{\mathscr{H},n}$ or $\hat{\lambda}_j = 0$. Then the
above bound and a Lidskii theorem for infinite-dimensional operators
\cite{kato87variatdiscrspect} yields
%
%
\begin{equation}
\label{eq30} \biggl( \sum_{j \geq1} (
\lambda_j - \hat{\lambda}_j)^{2}
\biggr)^{1/2} \leq2\sqrt{2} \sqrt{\frac{\tau}{n}}
\end{equation}
with probability at least $1 - 2e^{-\tau}$. For a given $d \geq1$
and $\tau> 0$, if the number $n$ of samples $X_i \sim F$ satisfies
\[
4\sqrt{2} \sqrt{\frac{\tau}{n}} < \lambda_{d} -
\lambda_{d+1},
\]
then with probability greater than $1 - 2e^{-\tau}$,
%
%
\begin{equation}
\label{eq32} \| \mathcal{P}_{d} - \hat{\mathcal{P}}_{d}
\|_{\mathrm{HS}} \leq\frac{2\sqrt{2} \sqrt{\tau}}{(\lambda_{d} - \lambda_{d+1})
\sqrt{n}},
\end{equation}
where $\mathcal{P}_{d}$ is the projection onto the subspace spanned
by the eigenfunctions corresponding to the $d$ largest eigenvalues
of $\mathscr{K}$, and $\hat{\mathcal{P}}_{d}$ is the projection onto
the subspace spanned by the eigenfunctions corresponding to the $d$
largest eigenvalues of $\mathscr{K}_{\mathscr{H},n}$.
\end{theorem}
\end{appendix}

\section*{Acknowledgements}

The authors thank the anonymous referees for their comments which have
improved the presentation and quality of the paper.


%

\printaddresses


\begin{thebibliography}{42}

\bibitem{Airoldi2008}
%
\begin{barticle}[author]
\bauthor{\bsnm{Airoldi},~\bfnm{E.~M.}\binits{E.~M.}},
\bauthor{\bsnm{Blei},~\bfnm{D.~M.}\binits{D.~M.}},
\bauthor{\bsnm{Fienberg},~\bfnm{S.~E.}\binits{S.~E.}} \AND
\bauthor{\bsnm{Xing},~\bfnm{E.~P.}\binits{E.~P.}}
(\byear{2008}).
\btitle{Mixed membership stochastic blockmodels}.
\bjournal{J. Mach. Learn. Res.}
\bvolume{9}
\bpages{1981--2014}.
\bptok{imsref}%
\end{barticle}
%
\endbibitem

\bibitem{bartlett06convex}
%
\begin{barticle}[mr]
\bauthor{\bsnm{Bartlett},~\bfnm{Peter~L.}\binits{P.~L.}},
\bauthor{\bsnm{Jordan},~\bfnm{Michael~I.}\binits{M.~I.}} \AND
\bauthor{\bsnm{McAuliffe},~\bfnm{Jon~D.}\binits{J.~D.}}
(\byear{2006}).
\btitle{Convexity, classification, and risk bounds}.
\bjournal{J. Amer. Statist. Assoc.}
\bvolume{101}
\bpages{138--156}.
\bid{doi={10.1198/016214505000000907}, issn={0162-1459}, mr={2268032}}
\bptok{imsref}%
\end{barticle}
%
\endbibitem

\bibitem{belkin05towarlaplac}
%
\begin{binproceedings}[mr]
\bauthor{\bsnm{Belkin},~\bfnm{Mikhail}\binits{M.}} \AND
\bauthor{\bsnm{Niyogi},~\bfnm{Partha}\binits{P.}}
(\byear{2005}).
\btitle{Towards a theoretical foundation for {L}aplacian-based manifold
methods}.
In \bbooktitle{Proceedings of the 18th Conference on Learning Theory (COLT
2005)}
\bpages{486--500}.
\bpublisher{Springer}, \blocation{Berlin}.
\bptok{imsref}%
\end{binproceedings}
%
\endbibitem

\bibitem{bengio04learnpca}
%
\begin{barticle}[author]
\bauthor{\bsnm{Bengio},~\bfnm{Y.}\binits{Y.}},
\bauthor{\bsnm{Vincent},~\bfnm{P.}\binits{P.}},
\bauthor{\bsnm{Paiement},~\bfnm{J.~F.}\binits{J.~F.}},
\bauthor{\bsnm{Delalleau},~\bfnm{O.}\binits{O.}},
\bauthor{\bsnm{Ouimet},~\bfnm{M.}\binits{M.}} \AND
\bauthor{\bsnm{Roux},~\bfnm{N.~Le}\binits{N.~L.}}
(\byear{2004}).
\btitle{Learning eigenfunctions links spectral embedding and kernel {PCA}}.
\bjournal{Neural Comput.}
\bvolume{16}
\bpages{2197--2219}.
\bptok{imsref}%
\end{barticle}
%
\endbibitem

\bibitem{biau05functhilberspaces}
%
\begin{barticle}[mr]
\bauthor{\bsnm{Biau},~\bfnm{G{\'e}rard}\binits{G.}},
\bauthor{\bsnm{Bunea},~\bfnm{Florentina}\binits{F.}} \AND
\bauthor{\bsnm{Wegkamp},~\bfnm{Marten~H.}\binits{M.~H.}}
(\byear{2005}).
\btitle{Functional classification in {H}ilbert spaces}.
\bjournal{IEEE Trans. Inform. Theory}
\bvolume{51}
\bpages{2163--2172}.
\bid{doi={10.1109/TIT.2005.847705}, issn={0018-9448}, mr={2235289}}
\bptok{imsref}%
\end{barticle}
%
\endbibitem

\bibitem{blanchard08statisperfor}
%
\begin{barticle}[mr]
\bauthor{\bsnm{Blanchard},~\bfnm{Gilles}\binits{G.}},
\bauthor{\bsnm{Bousquet},~\bfnm{Olivier}\binits{O.}} \AND
\bauthor{\bsnm{Massart},~\bfnm{Pascal}\binits{P.}}
(\byear{2008}).
\btitle{Statistical performance of support vector machines}.
\bjournal{Ann. Statist.}
\bvolume{36}
\bpages{489--531}.
\bid{doi={10.1214/009053607000000839}, issn={0090-5364}, mr={2396805}}
\bptok{imsref}%
\end{barticle}
%
\endbibitem

\bibitem{blanchard07statis}
%
\begin{barticle}[author]
\bauthor{\bsnm{Blanchard},~\bfnm{G.}\binits{G.}},
\bauthor{\bsnm{Bousquet},~\bfnm{O.}\binits{O.}} \AND
\bauthor{\bsnm{Zwald},~\bfnm{L.}\binits{L.}}
(\byear{2007}).
\btitle{Statistical properties of kernel principal component analysis}.
\bjournal{Machine Learning}
\bvolume{66}
\bpages{259--294}.
\bptok{imsref}%
\end{barticle}
%
\endbibitem

\bibitem{bollobas07}
%
\begin{barticle}[mr]
\bauthor{\bsnm{Bollob{\'a}s},~\bfnm{B{\'e}la}\binits{B.}},
\bauthor{\bsnm{Janson},~\bfnm{Svante}\binits{S.}} \AND
\bauthor{\bsnm{Riordan},~\bfnm{Oliver}\binits{O.}}
(\byear{2007}).
\btitle{The phase transition in inhomogeneous random graphs}.
\bjournal{Random Structures Algorithms}
\bvolume{31}
\bpages{3--122}.
\bid{doi={10.1002/rsa.20168}, issn={1042-9832}, mr={2337396}}
\bptok{imsref}%
\end{barticle}
%
\endbibitem

\bibitem{braun06accurerrorboundeigenkernelmatrix}
%
\begin{barticle}[mr]
\bauthor{\bsnm{Braun},~\bfnm{Mikio~L.}\binits{M.~L.}}
(\byear{2006}).
\btitle{Accurate error bounds for the eigenvalues of the kernel matrix}.
\bjournal{J. Mach. Learn. Res.}
\bvolume{7}
\bpages{2303--2328}.
\bid{issn={1532-4435}, mr={2274441}}
\bptok{imsref}%
\end{barticle}
%
\endbibitem

\bibitem{chatterjee12matrixuniversingulvaluethres}
%
\begin{bmisc}[author]
\bauthor{\bsnm{Chatterjee},~\bfnm{S.}\binits{S.}}
(\byear{2012}).
\bhowpublished{Matrix estimation by universal singular value thresholding.
Preprint. Available at \url{http://arxiv.org/abs/1212.1247}}.
\bptok{imsref}%
\end{bmisc}
%
\endbibitem

\bibitem{chaudhuri12spect}
%
\begin{binproceedings}[author]
\bauthor{\bsnm{Chaudhuri},~\bfnm{K.}\binits{K.}},
\bauthor{\bsnm{Chung},~\bfnm{F.}\binits{F.}} \AND
\bauthor{\bsnm{Tsiatas},~\bfnm{A.}\binits{A.}}
(\byear{2012}).
\btitle{Spectral partitioning of graphs with general degrees and the extended
planted partition model}.
In \bbooktitle{Proceedings of the 25th Conference on Learning Theory}.
\bpublisher{Springer}, \blocation{Berlin}.
\bptok{imsref}%
\end{binproceedings}
%
\endbibitem

\bibitem{cucker12}
%
\begin{barticle}[mr]
\bauthor{\bsnm{Cucker},~\bfnm{Felipe}\binits{F.}} \AND
\bauthor{\bsnm{Smale},~\bfnm{Steve}\binits{S.}}
(\byear{2002}).
\btitle{On the mathematical foundations of learning}.
\bjournal{Bull. Amer. Math. Soc. (N.S.)}
\bvolume{39}
\bpages{1--49 (electronic)}.
\bid{doi={10.1090/S0273-0979-01-00923-5}, issn={0273-0979}, mr={1864085}}
\bptok{imsref}%
\end{barticle}
%
\endbibitem

\bibitem{davis70}
%
\begin{barticle}[mr]
\bauthor{\bsnm{Davis},~\bfnm{Chandler}\binits{C.}} \AND
\bauthor{\bsnm{Kahan},~\bfnm{W.~M.}\binits{W.~M.}}
(\byear{1970}).
\btitle{The rotation of eigenvectors by a perturbation. {III}}.
\bjournal{SIAM J. Numer. Anal.}
\bvolume{7}
\bpages{1--46}.
\bid{issn={0036-1429}, mr={0264450}}
\bptok{imsref}%
\end{barticle}
%
\endbibitem

\bibitem{devroye1996probabilistic}
%
\begin{bbook}[mr]
\bauthor{\bsnm{Devroye},~\bfnm{Luc}\binits{L.}},
\bauthor{\bsnm{Gy{\"o}rfi},~\bfnm{L{\'a}szl{\'o}}\binits{L.}} \AND
\bauthor{\bsnm{Lugosi},~\bfnm{G{\'a}bor}\binits{G.}}
(\byear{1996}).
\btitle{A Probabilistic Theory of Pattern Recognition}.
\bseries{Applications of Mathematics (New York)}
\bvolume{31}.
\bpublisher{Springer}, \blocation{New York}.
\bid{mr={1383093}}
\bptok{imsref}%
\end{bbook}
%
\endbibitem

\bibitem{diaconis08graphlimitexchanrandomgraph}
%
\begin{barticle}[mr]
\bauthor{\bsnm{Diaconis},~\bfnm{Persi}\binits{P.}} \AND
\bauthor{\bsnm{Janson},~\bfnm{Svante}\binits{S.}}
(\byear{2008}).
\btitle{Graph limits and exchangeable random graphs}.
\bjournal{Rend. Mat. Appl. (7)}
\bvolume{28}
\bpages{33--61}.
\bid{issn={1120-7183}, mr={2463439}}
\bptok{imsref}%
\end{barticle}
%
\endbibitem

\bibitem{fishkind2012consistent}
%
\begin{barticle}[mr]
\bauthor{\bsnm{Fishkind},~\bfnm{Donniell~E.}\binits{D.~E.}},
\bauthor{\bsnm{Sussman},~\bfnm{Daniel~L.}\binits{D.~L.}},
\bauthor{\bsnm{Tang},~\bfnm{Minh}\binits{M.}},
\bauthor{\bsnm{Vogelstein},~\bfnm{Joshua~T.}\binits{J.~T.}} \AND
\bauthor{\bsnm{Priebe},~\bfnm{Carey~E.}\binits{C.~E.}}
(\byear{2013}).
\btitle{Consistent adjacency-spectral partitioning for the stochastic
block model when the model parameters are unknown}.
\bjournal{SIAM J. Matrix Anal. Appl.}
\bvolume{34}
\bpages{23--39}.
\bid{doi={10.1137/120875600}, issn={0895-4798}, mr={3032990}}
\bptok{imsref}%
\end{barticle}
%
\endbibitem

\bibitem{hein5fromlaplac}
%
\begin{binproceedings}[mr]
\bauthor{\bsnm{Hein},~\bfnm{Matthias}\binits{M.}},
\bauthor{\bsnm{Audibert},~\bfnm{Jean-Yves}\binits{J.-Y.}} \AND
\bauthor{\bparticle{von} \bsnm{Luxburg},~\bfnm{Ulrike}\binits{U.}}
(\byear{2005}).
\btitle{From graphs to manifolds---Weak and strong pointwise
consistency of
graph {L}aplacians}.
In \bbooktitle{Proceedings of the 18th Conference on Learning Theory (COLT
2005)}
\bpages{470--485}.
\bpublisher{Springer}, \blocation{Berlin}.
\bptok{imsref}%
\end{binproceedings}
%
\endbibitem

\bibitem{hein07converlaplac}
%
\begin{barticle}[author]
\bauthor{\bsnm{Hein},~\bfnm{M.}\binits{M.}},
\bauthor{\bsnm{Audibert},~\bfnm{J.~Y.}\binits{J.~Y.}} \AND
\bauthor{\bparticle{von} \bsnm{Luxburg},~\bfnm{U.}\binits{U.}}
(\byear{2007}).
\btitle{Convergence of graph {L}aplacians on random neighbourhood graphs}.
\bjournal{J. Mach. Learn. Res.}
\bvolume{8}
\bpages{1325--1370}.
\bptok{imsref}%
\end{barticle}
%
\endbibitem

\bibitem{Hoff2002}
%
\begin{barticle}[mr]
\bauthor{\bsnm{Hoff},~\bfnm{Peter~D.}\binits{P.~D.}},
\bauthor{\bsnm{Raftery},~\bfnm{Adrian~E.}\binits{A.~E.}} \AND
\bauthor{\bsnm{Handcock},~\bfnm{Mark~S.}\binits{M.~S.}}
(\byear{2002}).
\btitle{Latent space approaches to social network analysis}.
\bjournal{J. Amer. Statist. Assoc.}
\bvolume{97}
\bpages{1090--1098}.
\bid{doi={10.1198/016214502388618906}, issn={0162-1459}, mr={1951262}}
\bptok{imsref}%
\end{barticle}
%
\endbibitem

\bibitem{Holland1983}
%
\begin{barticle}[mr]
\bauthor{\bsnm{Holland},~\bfnm{Paul~W.}\binits{P.~W.}},
\bauthor{\bsnm{Laskey},~\bfnm{Kathryn~Blackmond}\binits{K.~B.}} \AND
\bauthor{\bsnm{Leinhardt},~\bfnm{Samuel}\binits{S.}}
(\byear{1983}).
\btitle{Stochastic blockmodels: First steps}.
\bjournal{Social Networks}
\bvolume{5}
\bpages{109--137}.
\bid{doi={10.1016/0378-8733(83)90021-7}, issn={0378-8733}, mr={0718088}}
\bptok{imsref}%
\end{barticle}
%
\endbibitem

\bibitem{karrer11stoch}
%
\begin{barticle}[mr]
\bauthor{\bsnm{Karrer},~\bfnm{Brian}\binits{B.}} \AND
\bauthor{\bsnm{Newman},~\bfnm{M.~E.~J.}\binits{M.~E.~J.}}
(\byear{2011}).
\btitle{Stochastic blockmodels and community structure in networks}.
\bjournal{Phys. Rev. E (3)}
\bvolume{83}
\bpages{016107, 10}.
\bid{doi={10.1103/PhysRevE.83.016107}, issn={1539-3755}, mr={2788206}}
\bptok{imsref}%
\end{barticle}
%
\endbibitem

\bibitem{kato87variatdiscrspect}
%
\begin{barticle}[mr]
\bauthor{\bsnm{Kato},~\bfnm{Tosio}\binits{T.}}
(\byear{1987}).
\btitle{Variation of discrete spectra}.
\bjournal{Comm. Math. Phys.}
\bvolume{111}
\bpages{501--504}.
\bid{issn={0010-3616}, mr={0900507}}
\bptok{imsref}%
\end{barticle}
%
\endbibitem

\bibitem{koltchinskii00random}
%
\begin{barticle}[mr]
\bauthor{\bsnm{Koltchinskii},~\bfnm{Vladimir}\binits{V.}} \AND
\bauthor{\bsnm{Gin{\'e}},~\bfnm{Evarist}\binits{E.}}
(\byear{2000}).
\btitle{Random matrix approximation of spectra of integral operators}.
\bjournal{Bernoulli}
\bvolume{6}
\bpages{113--167}.
\bid{doi={10.2307/3318636}, issn={1350-7265}, mr={1781185}}
\bptok{imsref}%
\end{barticle}
%
\endbibitem

\bibitem{lugosi04b}
%
\begin{barticle}[mr]
\bauthor{\bsnm{Lugosi},~\bfnm{G{\'a}bor}\binits{G.}} \AND
\bauthor{\bsnm{Vayatis},~\bfnm{Nicolas}\binits{N.}}
(\byear{2004}).
\btitle{On the {B}ayes-risk consistency of regularized boosting methods}.
\bjournal{Ann. Statist.}
\bvolume{32}
\bpages{30--55}.
\bid{issn={0090-5364}, mr={2051000}}
\bptok{imsref}%
\end{barticle}
%
\endbibitem

\bibitem{micchelli06univer}
%
\begin{barticle}[mr]
\bauthor{\bsnm{Micchelli},~\bfnm{Charles~A.}\binits{C.~A.}},
\bauthor{\bsnm{Xu},~\bfnm{Yuesheng}\binits{Y.}} \AND
\bauthor{\bsnm{Zhang},~\bfnm{Haizhang}\binits{H.}}
(\byear{2006}).
\btitle{Universal kernels}.
\bjournal{J. Mach. Learn. Res.}
\bvolume{7}
\bpages{2651--2667}.
\bid{issn={1532-4435}, mr={2274454}}
\bptok{imsref}%
\end{barticle}
%
\endbibitem

\bibitem{oliveira2010concentration}
%
\begin{bmisc}[author]
\bauthor{\bsnm{Oliveira},~\bfnm{R.~I.}\binits{R.~I.}}
(\byear{2010}).
\bhowpublished{Concentration of the adjacency matrix and of the
Laplacian in random
graphs with independent edges.
Preprint. Available at \url{http://arxiv.org/abs/0911.0600}}.
\bptok{imsref}%
\end{bmisc}
%
\endbibitem

\bibitem{rohe2011spectral}
%
\begin{barticle}[mr]
\bauthor{\bsnm{Rohe},~\bfnm{Karl}\binits{K.}},
\bauthor{\bsnm{Chatterjee},~\bfnm{Sourav}\binits{S.}} \AND
\bauthor{\bsnm{Yu},~\bfnm{Bin}\binits{B.}}
(\byear{2011}).
\btitle{Spectral clustering and the high-dimensional stochastic blockmodel}.
\bjournal{Ann. Statist.}
\bvolume{39}
\bpages{1878--1915}.
\bid{doi={10.1214/11-AOS887}, issn={0090-5364}, mr={2893856}}
\bptok{imsref}%
\end{barticle}
%
\endbibitem

\bibitem{rosasco10integoperat}
%
\begin{barticle}[mr]
\bauthor{\bsnm{Rosasco},~\bfnm{Lorenzo}\binits{L.}},
\bauthor{\bsnm{Belkin},~\bfnm{Mikhail}\binits{M.}} \AND
\bauthor{\bsnm{De~Vito},~\bfnm{Ernesto}\binits{E.}}
(\byear{2010}).
\btitle{On learning with integral operators}.
\bjournal{J.~Mach. Learn. Res.}
\bvolume{11}
\bpages{905--934}.
\bid{issn={1532-4435}, mr={2600634}}
\bptok{imsref}%
\end{barticle}
%
\endbibitem

\bibitem{shawe-taylor05eigengramgenerpca}
%
\begin{barticle}[mr]
\bauthor{\bsnm{Shawe-Taylor},~\bfnm{John}\binits{J.}},
\bauthor{\bsnm{Williams},~\bfnm{Christopher K.~I.}\binits{C.~K.~I.}},
\bauthor{\bsnm{Cristianini},~\bfnm{Nello}\binits{N.}} \AND
\bauthor{\bsnm{Kandola},~\bfnm{Jaz}\binits{J.}}
(\byear{2005}).
\btitle{On the eigenspectrum of the {G}ram matrix and the
generalization error
of kernel-{PCA}}.
\bjournal{IEEE Trans. Inform. Theory}
\bvolume{51}
\bpages{2510--2522}.
\bid{doi={10.1109/TIT.2005.850052}, issn={0018-9448}, mr={2246374}}
\bptok{imsref}%
\end{barticle}
%
\endbibitem

\bibitem{sriperumbudur11univercharackernelrkhsembedmeasur}
%
\begin{barticle}[mr]
\bauthor{\bsnm{Sriperumbudur},~\bfnm{Bharath~K.}\binits{B.~K.}},
\bauthor{\bsnm{Fukumizu},~\bfnm{Kenji}\binits{K.}} \AND
\bauthor{\bsnm{Lanckriet},~\bfnm{Gert R.~G.}\binits{G.~R.~G.}}
(\byear{2011}).
\btitle{Universality, characteristic kernels and {RKHS} embedding of measures}.
\bjournal{J. Mach. Learn. Res.}
\bvolume{12}
\bpages{2389--2410}.
\bid{issn={1532-4435}, mr={2825431}}
\bptok{imsref}%
\end{barticle}
%
\endbibitem

\bibitem{steinwart01supporvectormachin}
%
\begin{barticle}[mr]
\bauthor{\bsnm{Steinwart},~\bfnm{Ingo}\binits{I.}}
(\byear{2002}).
\btitle{On the influence of the kernel on the consistency of support vector
machines}.
\bjournal{J. Mach. Learn. Res.}
\bvolume{2}
\bpages{67--93}.
\bid{doi={10.1162/153244302760185252}, issn={1532-4435}, mr={1883281}}
\bptnote{check year}%
\bptok{imsref}%
\end{barticle}
%
\endbibitem

\bibitem{steinwart02suppor}
%
\begin{barticle}[mr]
\bauthor{\bsnm{Steinwart},~\bfnm{Ingo}\binits{I.}}
(\byear{2002}).
\btitle{Support vector machines are universally consistent}.
\bjournal{J. Complexity}
\bvolume{18}
\bpages{768--791}.
\bid{doi={10.1006/jcom.2002.0642}, issn={0885-064X}, mr={1928806}}
\bptok{imsref}%
\end{barticle}
%
\endbibitem

\bibitem{steinwart12mercer}
%
\begin{barticle}[mr]
\bauthor{\bsnm{Steinwart},~\bfnm{Ingo}\binits{I.}} \AND
\bauthor{\bsnm{Scovel},~\bfnm{Clint}\binits{C.}}
(\byear{2012}).
\btitle{Mercer's theorem on general domains: On the interaction between
measures, kernels, and {RKHS}s}.
\bjournal{Constr. Approx.}
\bvolume{35}
\bpages{363--417}.
\bid{doi={10.1007/s00365-012-9153-3}, issn={0176-4276}, mr={2914365}}
\bptok{imsref}%
\end{barticle}
%
\endbibitem

\bibitem{stone1977consistent}
%
\begin{barticle}[mr]
\bauthor{\bsnm{Stone},~\bfnm{Charles~J.}\binits{C.~J.}}
(\byear{1977}).
\btitle{Consistent nonparametric regression}.
\bjournal{Ann. Statist.}
\bvolume{5}
\bpages{595--620}.
\bptok{imsref}%
\end{barticle}
%
\endbibitem

\bibitem{sussman12}
%
\begin{barticle}[mr]
\bauthor{\bsnm{Sussman},~\bfnm{Daniel~L.}\binits{D.~L.}},
\bauthor{\bsnm{Tang},~\bfnm{Minh}\binits{M.}},
\bauthor{\bsnm{Fishkind},~\bfnm{Donniell~E.}\binits{D.~E.}} \AND
\bauthor{\bsnm{Priebe},~\bfnm{Carey~E.}\binits{C.~E.}}
(\byear{2012}).
\btitle{A consistent adjacency spectral embedding for stochastic blockmodel
graphs}.
\bjournal{J. Amer. Statist. Assoc.}
\bvolume{107}
\bpages{1119--1128}.
\bid{doi={10.1080/01621459.2012.699795}, issn={0162-1459}, mr={3010899}}
\bptok{imsref}%
\end{barticle}
%
\endbibitem

\bibitem{sussman12univer}
%
\begin{bmisc}[author]
\bauthor{\bsnm{Sussman},~\bfnm{D.~L.}\binits{D.~L.}},
\bauthor{\bsnm{Tang},~\bfnm{M.}\binits{M.}} \AND
\bauthor{\bsnm{Priebe},~\bfnm{C.~E.}\binits{C.~E.}}
(\byear{2012}).
\bhowpublished{Consistent latent position estimation and vertex
classification for
random dot product graphs.
Preprint. Available at \url{http://arxiv.org/abs/1207.6745}}.
\bptok{imsref}%
\end{bmisc}
%
\endbibitem

\bibitem{tropp11freed}
%
\begin{barticle}[mr]
\bauthor{\bsnm{Tropp},~\bfnm{Joel~A.}\binits{J.~A.}}
(\byear{2011}).
\btitle{Freedman's inequality for matrix martingales}.
\bjournal{Electron. Commun. Probab.}
\bvolume{16}
\bpages{262--270}.
\bid{doi={10.1214/ECP.v16-1624}, issn={1083-589X}, mr={2802042}}
\bptok{imsref}%
\end{barticle}
%
\endbibitem

\bibitem{luxburg08consis}
%
\begin{barticle}[mr]
\bauthor{\bparticle{von} \bsnm{Luxburg},~\bfnm{Ulrike}\binits{U.}},
\bauthor{\bsnm{Belkin},~\bfnm{Mikhail}\binits{M.}} \AND
\bauthor{\bsnm{Bousquet},~\bfnm{Olivier}\binits{O.}}
(\byear{2008}).
\btitle{Consistency of spectral clustering}.
\bjournal{Ann. Statist.}
\bvolume{36}
\bpages{555--586}.
\bid{doi={10.1214/009053607000000640}, issn={0090-5364}, mr={2396807}}
\bptok{imsref}%
\end{barticle}
%
\endbibitem

\bibitem{young2007random}
%
\begin{bincollection}[mr]
\bauthor{\bsnm{Young},~\bfnm{Stephen~J.}\binits{S.~J.}} \AND
\bauthor{\bsnm{Scheinerman},~\bfnm{Edward~R.}\binits{E.~R.}}
(\byear{2007}).
\btitle{Random dot product graph models for social networks}.
In \bbooktitle{Algorithms and Models for the Web-Graph}.
\bseries{Lecture Notes in Computer Science}
\bvolume{4863}
\bpages{138--149}.
\bpublisher{Springer}, \blocation{Berlin}.
\bid{doi={10.1007/978-3-540-77004-6_11}, mr={2504912}}
\bptok{imsref}%
\end{bincollection}
%
\endbibitem

\bibitem{zhang04statis}
%
\begin{barticle}[mr]
\bauthor{\bsnm{Zhang},~\bfnm{Tong}\binits{T.}}
(\byear{2004}).
\btitle{Statistical behavior and consistency of classification methods
based on
convex risk minimization}.
\bjournal{Ann. Statist.}
\bvolume{32}
\bpages{56--85}.
\bid{doi={10.1214/aos/1079120130}, issn={0090-5364}, mr={2051001}}
\bptok{imsref}%
\end{barticle}
%
\endbibitem

\bibitem{zwald06}
%
\begin{binproceedings}[author]
\bauthor{\bsnm{Zwald},~\bfnm{L.}\binits{L.}} \AND
\bauthor{\bsnm{Blanchard},~\bfnm{G.}\binits{G.}}
(\byear{2006}).
\btitle{On the convergence of eigenspaces in kernel principal components
analysis}.
In \bbooktitle{Advances in Neural Information Processing Systems (NIPS 05)}
\bvolume{18}
\bpages{1649--1656}.
\bpublisher{MIT Press}, \blocation{Cambridge}.
\bptok{imsref}%
\end{binproceedings}
%
\endbibitem

\end{thebibliography}
\end{document}